\documentclass[journal]{IEEEtran}
\usepackage{threeparttable}
\usepackage{graphicx}
\usepackage{amsmath}
\usepackage{amsfonts}
\usepackage{multirow}
\usepackage{multicol}
\usepackage{cite}
\usepackage[numbers]{natbib}
\usepackage{setspace}
\usepackage{color}
\ifCLASSINFOpdf

\else

\fi

\begin{document}

\title{BNAS: An Efficient Neural Architecture Search Approach Using Broad Scalable Architecture}

\author{Zixiang~Ding,
        Yaran~Chen,
        Nannan~Li,
        Dongbin~Zhao,~\IEEEmembership{Fellow,~IEEE},
        Zhiquan~Sun,
        and~C.L.Philip~Chen,~\IEEEmembership{Fellow,~IEEE}
\thanks{Z. Ding, Y. Chen, N. Li and D. Zhao are with the State Key Laboratory of Management and Control for Complex Systems, Institute of Automation, Chinese Academy of Sciences, Beijing 100190, China, and also with the University of Chinese Academy of Sciences, Beijing 100049, China (email : \{dingzixiang2018, chenyaran2013, linannan2017, dongbin.zhao\}@ia.ac.cn).}
\thanks{Z. Sun is with the School of Automation and Electric Engineering, University of Science and Technology Beijing, Beijing 100083, China (email : 41723308@xs.ustb.edu.cn).}
\thanks{C. L. P. Chen is with the School of Computer Science \& Engineering, South China University of Technology, Guangzhou, Guangdong 510006, China, and also with the College of Navigation, Dalian Maritime University, Dalian 116026, China (e-mail: philip.chen@ieee.org).}

\thanks{This work was supported by No. GJHZ1849 International Partnership Program of Chinese Academy of Sciences.}}



\maketitle

\begin{abstract}
Efficient Neural Architecture Search (ENAS) achieves novel efficiency for learning architecture with high-performance via parameter sharing and reinforcement learning. In the phase of architecture search, ENAS employs deep scalable architecture as search space whose training process consumes most of search cost. Moreover, time consuming of model training is proportional to the depth of deep scalable architecture. Through experiments using ENAS on CIFAR-10, we find that layer reduction of scalable architecture is an effective way to accelerate the search process of ENAS but suffers from prohibitive performance drop in the phase of architecture estimation.

In this paper, we propose Broad Neural Architecture Search (BNAS) where we elaborately design broad scalable architecture dubbed Broad Convolutional Neural Network (BCNN) to solve the above issue. On one hand, the proposed broad scalable architecture has fast training speed due to its shallow topology. Moreover, we also adopt reinforcement learning and parameter sharing used in ENAS as the optimization strategy of BNAS. Hence, the proposed approach can achieve higher search efficiency. On the other hand, the broad scalable architecture extracts multi-scale features and enhancement representations, and feeds them into global average pooling layer to yield more reasonable and comprehensive representations. Therefore, the performance of broad scalable architecture can be promised. In particular, we also develop two variants for BNAS who modify the topology of BCNN. In order to verify the effectiveness of BNAS, several experiments are performed and experimental results show that 1) BNAS delivers 0.19 days which is 2.37x less expensive than ENAS who ranks the best in reinforcement learning-based NAS approaches, 2) compared with small-size (0.5 millions parameters) and medium-size (1.1 millions parameters) models, the architecture learned by BNAS obtains state-of-the-art performance (3.58\% and 3.24\% test error) on CIFAR-10, 3) the learned architecture achieves 25.3\% top-1 error on ImageNet just using 3.9 millions parameters.
\end{abstract}

\section{Introduction}
\label{introduction}


As a biologically inspired deep learning technique, convolutional neural networks (CNNs) have powerful ability to solve many intractable issues, e.g. computer vision tasks \cite{zhao2017deep,chen2018multi,liu2020stroke,mellouli2019morphological}, game artificial intelligence \cite{shao2019starcraft,shao2018learning}, intelligent transportation system \cite{li2018deepsign} and intelligent robot \cite{li2019deep}. However, its remarkable performance mainly depends on human experts with a considerable amount of expertise. Recently, Neural Architecture Search (NAS) \cite{zoph2017neural} which automates the process of model design has gained ground in recent several years. Computer vision tasks (e.g. image classification \cite{brock2017smash,zoph2018learning,chen2020modulenet,sun2019completely}, semantic segmentation \cite{liu2019auto}) and other artificial intelligence related tasks (e.g. natural language processing \cite{liu2018darts,pham2018efficient,zoph2017neural}) can all be solved by NAS with surprising performance. However, early approaches \cite{zoph2017neural,zoph2018learning,real2018regularized} suffered from the issue of inefficiency. To solve this issue, some one-shot
approaches \cite{brock2017smash,pham2018efficient,ding2019simplified,liu2018darts,li2019light,li2020shift} were proposed. Generally speaking, one-shot NAS approaches sample Cells (Normal Cell and Reduction Cell), micro search space presented in \cite{zoph2018learning}, from a family of predefined candidate operations depending on a policy. The sampled Cells are treated as building block of deep scalable architecture, i.e. child model, whose performance is used for updating the parameters of policy. These one-shot approaches avoid retraining each candidate deep scalable architecture from scratch so that high efficiency can be promised.

In particular, Efficient Neural Architecture Search (ENAS) \cite{pham2018efficient} delivers the efficiency of 0.45 GPU days (ranks the best in reinforcement learning-based NAS frameworks) by sharing parameters of one-shot model where all possible child models are defined. In order to improve the performance and robustness of discovered architecture, ENAS sets the number of Cells to 8 (6 Normal Cells and 2 Reduction Cells) and 17 (15 Normal Cells and 2 Reduction Cells) in the phase of architecture search and architecture estimation, respectively. There is a loop in the architecture search phase: 1) ENAS trains one-shot model on proxy data set (e.g. CIFAR-10) in a single epoch where a recurrent neural network (RNN) controller is adopted to sample a child model from one-shot model in each step. 2) For learning better architecture, the RNN controller is optimized by reinforcement learning in fixed steps. Here, a child model represented by a sequence is sampled by the RNN controller. For reinforcement learning, the sequence is treated as a token of actions, and validation accuracy of child model is treated as reward. On one hand, time consuming of model training and inference is proportional to the depth of model. On the other hand, the topologies of one-shot model and child model are all deep in ENAS. As a result, training one-shot model and obtaining validation accuracy of child model consume most of search cost in ENAS. Fortunately, depth reduction of scalable architecture in architecture search phase can ameliorate the above issues.

\begin{figure}[!t]
\centering
\includegraphics[width=9.5cm]{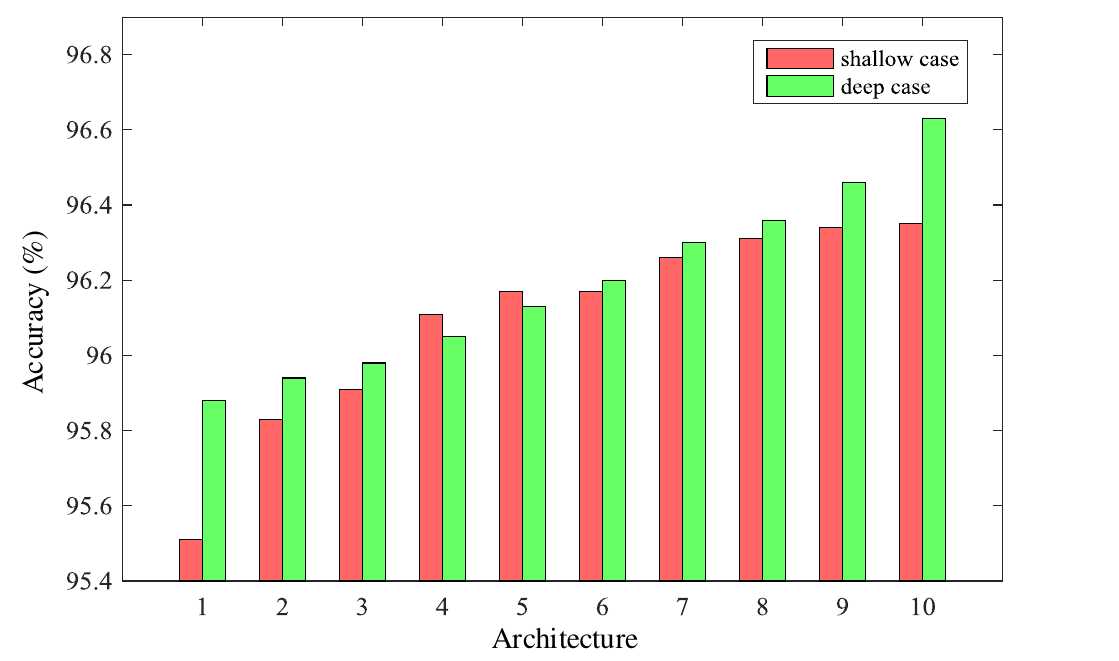}
\caption{Performance (ranks the worst from the best) of ten candidate architectures of ENAS when using various number of Cells of deep scalable architecture in search phase on CIFAR-10 without Cutout \cite{devries2017improved} technique. \emph{Shallow case}: Test accuracy of architectures using 5 Cells in search phase and 17 Cells in estimation phase. \emph{Deep case}: Test accuracy of architectures using 8 Cells in search phase and 17 Cells in estimation phase.}
\label{enas_difference}
\end{figure}

Experiments on ENAS$\footnote{https://github.com/melodyguan/enas}$ are carried out to compare the discrepancies with respect to efficiency and accuracy when using 5 (shallow case) and 8 (deep case, also default setting for ENAS) Cells in the architecture search phase of ENAS on CIFAR-10. Experimental results (see Fig. \ref{enas_difference}) show that depth reduction of scalable architecture is able to improve the search efficiency of ENAS, but suffers from prohibitive performance drop. Moreover, with a single GeForce GTX 2080Ti GPU, the shallow case and deep case cost 0.26 and 0.38 days, respectively. However, the deep case delivers better performance (about 0.3\% between the top performing architectures of two cases) than the shallow one. Hence, designing high-performance scalable architecture with shallow topology is an effective way to develop more efficient NAS approach.

In this paper, we propose Broad Neural Architecture Search (BNAS), an automatic architecture search approach with state-of-the-art efficiency. Different from other NAS approaches, in BNAS, an elaborately designed broad scalable architecture dubbed Broad Convolutional Neural Network (BCNN) instead of a deep one is discovered by parameter sharing and reinforcement learning. BCNN consists of convolution blocks and enhancement blocks. On one hand, each convolution block is fully-connected with enhancement blocks. On the other hand, all outputs of convolution blocks and enhancement blocks are fed into the global poling layer as inputs. Furthermore, we also develop two variants for BNAS by modifying the topology of broad scalable architecture: 1) Cascade of Convolution blocks with its Last block connected to the Enhancement blocks Broad Neural Architecture Search (BNAS-CCLE) and 2) Cascade of Convolution blocks and Enhancement blocks Broad Neural Architecture Search (BNAS-CCE). The proposed several broad scalable architectures extract multi-scale features and enhancement representations, and feed them into global average pooling layer to yield more reasonable and comprehensive representations, so that the performance of architectures learned by BNAS and its two variants can be promised. Our contributions can be summarized as follows:

\begin{itemize}
  \item We propose BNAS to further improve the efficiency of NAS by replacing the deep scalable architecture with a broad one dubbed BCNN who is elaborately designed for satisfactory performance and fast search efficiency simultaneously.
  \item We also propose two developed versions of BNAS who modify the topology of BCNN dubbed BNAS-CCLE and BNAS-CCE. All of BNAS and its two variants excel in discovering high-performance architecture with fast search efficiency.
  \item We achieve 2.37x less search cost (with a single GeForce GTX 1080Ti GPU on CIFAR-10 in 0.19 days) than ENAS \cite{pham2018efficient} who ranks the best in reinforcement learning-based NAS approaches. Furthermore, through extensive experiments on CIFAR-10, we show that the architecture learned by BNAS obtains state-of-the-art performance (3.58\% and 3.24\% test error) compared with small-size (0.5 millions parameters) and medium-size (1.1 millions parameters) models.
  \item We transfer the learned architecture for large scale image classification task to shed light to the powerful transferability and multi-scale features extraction capacity of BCNN. The learned architecture achieves 25.7\% top-1 error just using 3.9 millions parameters on ImageNet.
\end{itemize}

The remainder of this paper is organized as follows. In Section \ref{related work}, we review related work with respect to this paper. Then, the proposed BNAS is described in Section \ref{the proposed approach}. Subsequently, two variants of BNAS are given in Section \ref{variants}. Next, experiments on two data set are performed, and qualitative and quantitative analysis is given in Section \ref{experiments and analysis}. At last, we draw some conclusions in Section \ref{conclusions}.

\begin{figure*}[!t]
\centering
\includegraphics[width=16cm]{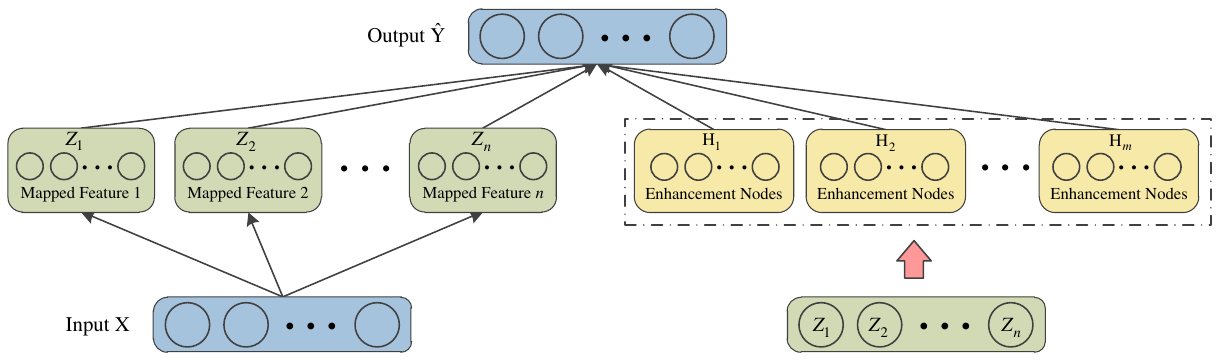}
\caption{Structure of BLS. BLS consisted of two types of groups, feature mapped nodes group (FMNG) and enhancement nodes group (ENG), where each one contained a group of neurons. The inputs of each FMNG and ENG were input data and the combination of outputs of every FMNG, respectively}
\label{bls}
\end{figure*}

\section{Related Work}
\label{related work}


\subsection{Neural Architecture Search}
\label{NAS}

From the point of view of optimization strategy, NAS approaches can be divided into three categories mainly: reinforcement learning (RL)-based \cite{zoph2017neural,zoph2018learning,pham2018efficient}, evolutionary algorithm (EA)-based \cite{real2018regularized,elsken2018efficient,dong2018dpp,zhu2019multi} and gradient-based \cite{liu2018darts,chen2019progressive,xu2019pc}. Below, we will introduce several significant NAS approaches with respect to various optimization strategies.

\begin{figure}[!t]
\centering
\includegraphics[width=9cm]{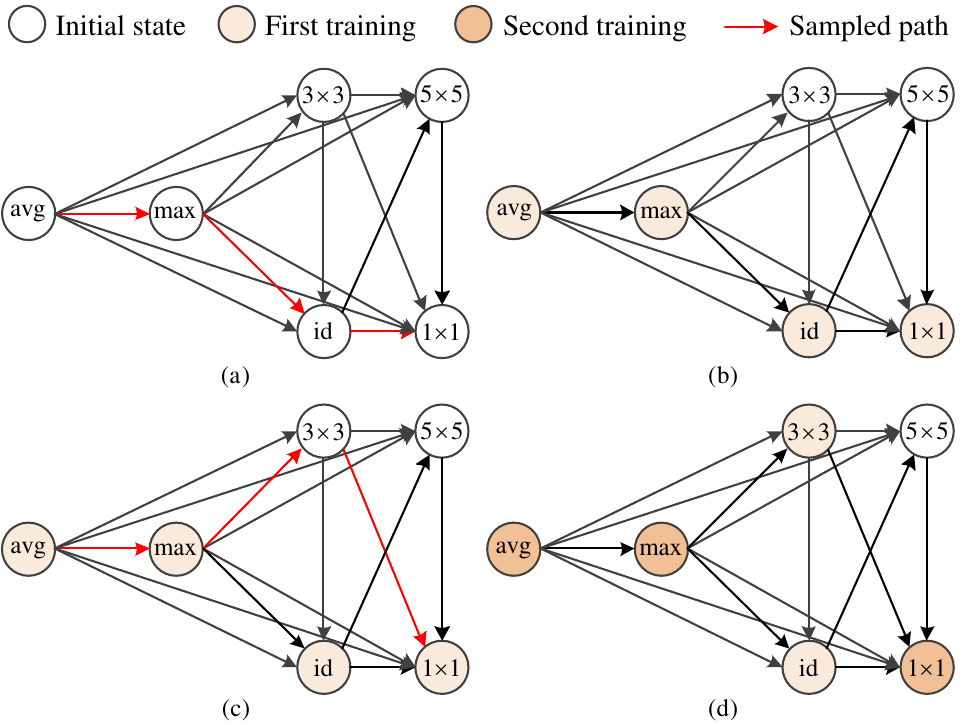}
\caption{Mechanism of parameters sharing. Cell was treated as a directed acyclic graph (DAG), and trained as follows. (a) Controller sampled a path in the DAG, (b) the sampled operations were trained with a mini-batch data, (c) controller sampled another path, and (d) another mini-batch was employed to train the sampled operations whose weights were inherited from previous training step rather than trained from scratch.}
\label{sharing}
\end{figure}

In the RL-based NAS framework, an architecture generator named controller is trained in a loop by policy gradient-based RL, where architecture sequence and validation accuracy are treated as a list of tokens of actions and reward, respectively. \citet{zoph2017neural} adopted REINFORCE to sample whole child model that was trained from scratch with considerable computational cost. A micro search space named Cell was proposed to improve the search efficiency of NAS in the following work dubbed NASNet \cite{zoph2018learning}, where child model consists of several Cells, and was trained with unacceptable computational resource (i.e. 1800 GPU days) due to the trained weights were thrown away yet. ENAS \cite{pham2018efficient} delivered 1000x faster search efficiency (i.e. 0.45 days with a single GeForce GTX 1080Ti GPU) than NASNet via parameters sharing shown in Fig. \ref{sharing}. All trained weights were inherited from previous training process rather than trained from scratch.

Furthermore, validation accuracy is treated as the fitness to guide the EA-based NAS approaches for finding better architecture. Moreover, crossover and variation are also employed for architecture evolution \cite{real2018regularized}. The combination of multi-objective evolutionary algorithm and Lamarckian inheritance mechanism was proposed to improve the performance of NAS in LEMONADE \cite{elsken2018efficient}. Pareto-optimal architectures were discovered for device-related and device-agnostic objectives in DPP-Net \cite{dong2018dpp}. Beyond that, the strategy of progressive search \cite{liu2018progressive} was employed in DPP-Net to accelerate its efficiency of 4 GPU days that was about 800x less expensive than AmoebaNet \cite{real2018regularized} using 3150 GPU days for architecture search.

Different from the previous two types of NAS, gradient-based framework transferred the architecture optimization issue from discrete space to continuous one. \citet{liu2018darts} proposed Differentiable ARchiTecture Search (DARTS). DARTS discovered the computation Cells within a continuous domain for formulating NAS in a differentiable way. DARTS achieved three order of magnitudes less expensive than previous approaches \cite{zoph2017neural,zoph2018learning}. Subsequently, several developed versions of DARTS (e.g., P-DARTS \cite{chen2019progressive}, PC-DARTS \cite{xu2019pc}) were proposed. Based on the framework of DARTS, P-DARTS modified the scalable architecture in a progressive way, and delivered novel search efficiency of 0.3 GPU days. PC-DARTS activated partial channel connections of Cell to reduce the memory usage of DARTS so that lager batch size could be set in the architecture search phase. Lager batch size contributed to reduce the search cost and uncertainty in NAS. Moreover, PC-DARTS achieved state-of-the-art efficiency (0.1 days) on a single GeForce GTX 1080Ti GPU.

\subsection{Broad Learning System}

BLS \cite{chen2017broad,chen2018universal} was proposed as a developed model of Random Vector Functional-Link Neural Network (RVFLNN) \cite{pao1992functional,pao1994learning} who took input data directly to build enhancement nodes. Different from RVFLNN, a set of mapped features were established in BLS by the input data firstly for achieving satisfactory performance.

We show the structure of BLS in Fig. \ref{bls}. Feature mapping nodes and enhancement nodes were two main components of BLS. In \cite{chen2017broad}, the mechanism of BLS was introduced: 1) Nonlinear transformation functions of feature mapping nodes were applied to generate the mapped features of input data. 2) The mapped features were enhanced to generate enhancement features by enhancement nodes with randomly generated weights. 3) All the mapped features and enhancement features were used to deliver the final result. Chen \emph{et al}. introduced several variants of BLS in \cite{chen2018universal}, e.g. Cascade of Convolution Feature mapping nodes Broad Learning System (CCFBLS), Cascade of Feature mapping nodes Broad Learning System (CFBLS), Cascade of Feature mapping nodes and Enhancement nodes Broad Learning System (CFEBLS). Below, CCFBLS, CFBLS and CFEBLS are introduced in details.

\begin{figure*}[!t]
\centering
\includegraphics[width=16.3cm]{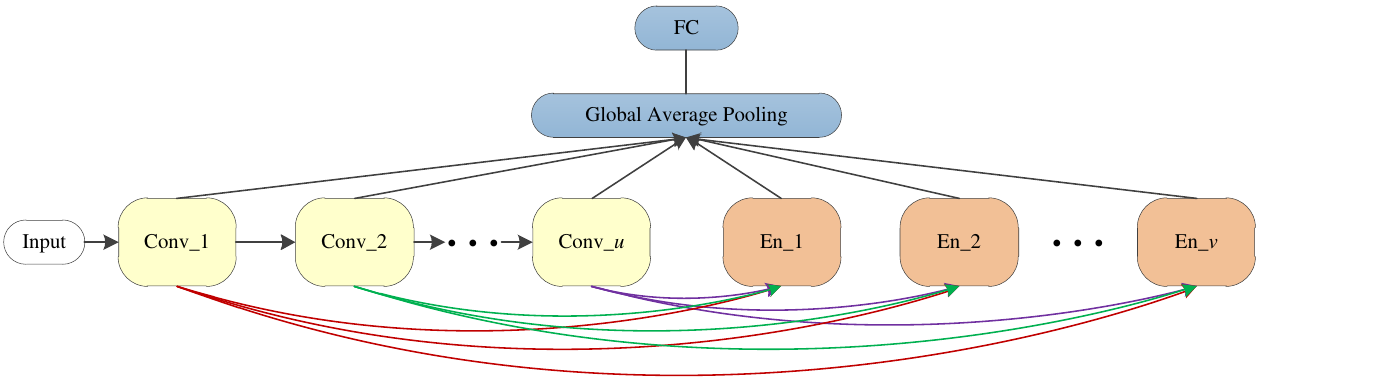}
\caption{The topology of broad scalable architecture dubbed BCNN employed in BNAS. The proposed BCNN stacks $u$ convolution blocks one after another, and feeds outputs of every convolution block into each enhancement block as the input for obtaining enhancement feature representations. The convolution and enhancement features from every convolution and enhancement block (i.e. multi-scale feature fusion) are all connected with the global average pooling layer to yield more reasonable and comprehensive representations for achieving promised performance of the proposed BCNN. Moreover, we also insert prior knowledge into 1) the connection between convolution and enhancement blocks, and 2) the connection between all blocks and global average pooling layer, to achieve satisfactory performance.}
\label{BCNN}
\end{figure*}

Feature mapping nodes and enhancement nodes made up CCFBLS. As described in \cite{chen2018universal}, the mapped features were generated by cascade of convolution and pooling operations in feature mapping nodes. Then, these mapped features were enhanced by a nonlinear activation function to obtain a series of enhancement features. Finally, all of the mapped features and enhancement features were connected directly with the desired output. CCFBLS was not only broad but also deep. As a result, CCFBLS could extract multi-layer features and representations which were more reasonable and comprehensive compared with other models only with deep structure. There were two main discrepancies among CCFBLS and other two variants: 1) The feature mapping nodes of CFBLS and CFEBLS were neurons rather than convolution and pooling operations. 2) Their topologies were different. For CFBLS, the output of each group of feature mapping nodes was fed into every enhancement nodes group as input. For CFEBLS, the first enhancement nodes group accepted those outputs from each feature mapping nodes group as input. Other enhancement nodes group treated the output of its previous one as input.

\section{Broad Neural Architecture Search}
\label{the proposed approach}

BNAS employs the combination of broad scalable architecture and parameters sharing to deliver faster search efficiency than ENAS who ranks the best in RL-based NAS approaches. First of all, we introduce the design details of the proposed broad scalable architecture dubbed BCNN who can solve the side-effect of layer reduction in Cell-based NAS approaches. Subsequently, the efficient optimization strategy of BNAS is given. At last, we show the differences between BLS and the proposed BCNN.

\subsection{Broad Convolutional Neural Network}
\label{bcnn}

As described in Section \ref{introduction}, we draw a conclusion through a set of comparative experiments that layer reduction is able to improve the search efficiency of ENAS but suffers from prohibitive performance drop. Inspired by BLS who achieves satisfactory performance using broad topology, we propose a broad scalable architecture dubbed BCNN who can deliver fast training and inference speed, to ameliorate the above issue. For intuitional understanding, the structure of BCNN is depicted in Fig. \ref{BCNN}. The proposed BCNN consists of $u$ convolution blocks denoted as $Conv\_i \ (i=1,2,\dots,u)$ and $v$ enhancement blocks denoted as $En\_j \ (j=1,2,\dots,v)$ which are used for feature extraction and enhancement, respectively. In the convolution block, there are $k+1$ convolution Cells: $k$ deep Cells and a single broad Cell which are utilized for deep and broad features extraction, respectively. Moreover, $u$ is determined by the size of input images. For example, we set $u=2$ for the experiments on CIFAR-10 with $32 \times 32$ pixels. The other two parameters $k$ and $v$ need to be defined by user. Through substantial experiments on CIFAR-10, we will analyze the influence with regard to $k$ and $v$ in Section \ref{Hyper-parameters}. In the following, we will introduce the details about four important parts of BCNN: convolution block, enhancement block, multi-scale feature fusion and prior knowledge embedding.

\begin{figure}[!ht]
\centering
\includegraphics[width=8.8cm]{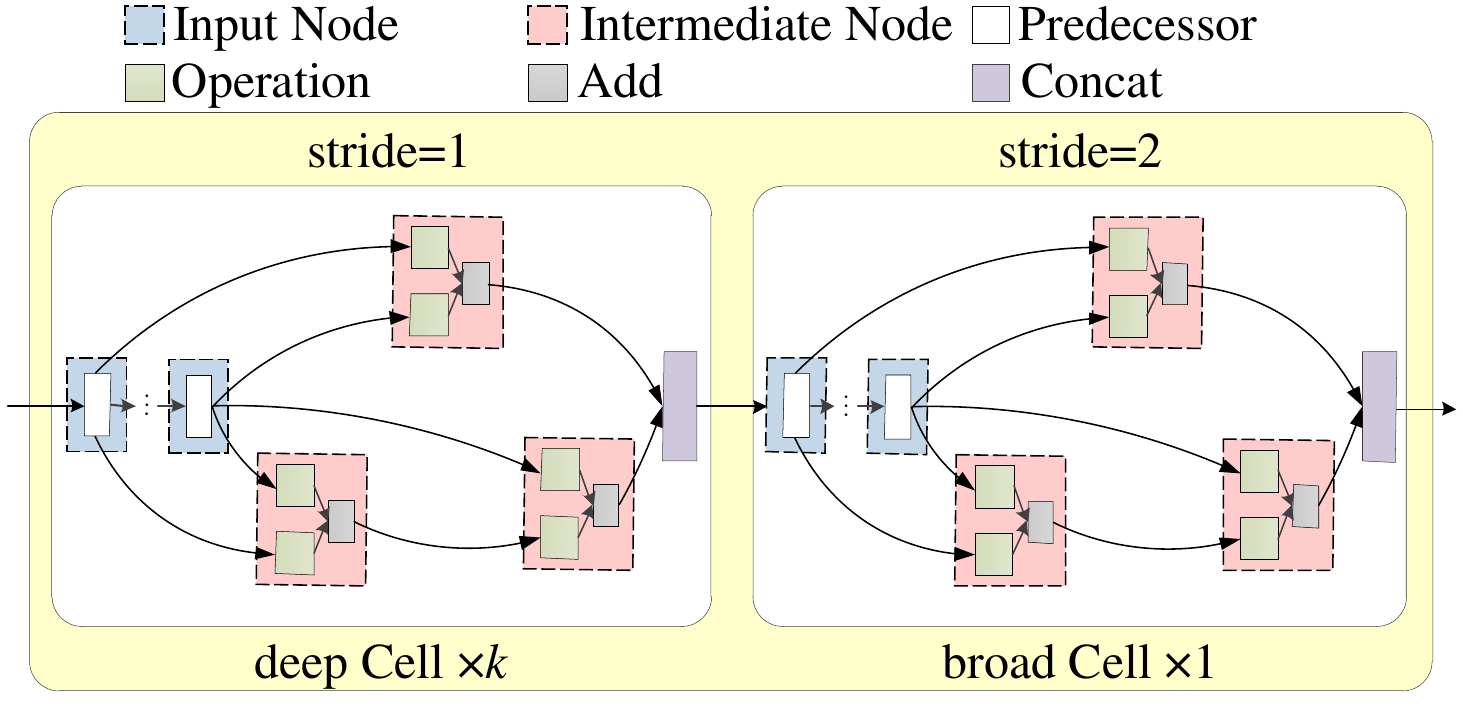}
\caption{The structure of convolution block.}
\label{convolution_block}
\end{figure}
\paragraph{Convolution Block} We visualize the structure of convolution block in Fig. \ref{convolution_block}. In each convolution block, the deep Cells and broad Cell have same topologies but various strides: one for the deep and two for the broad. In order to extract broad features from the output features of final deep Cell, the broad Cell returns the feature maps with half width, half height and double channels (i.e. broad features).
For convolution block $Conv\_i$, its deep feature mapping $\pmb{Z}^{(i)}_{h}(h=1,2,\dots, k)$ and broad feature mapping $\pmb{Z}^{(i)}_{k+1}$ can be defined as
\begin{align}
\pmb{Z}^{(i)}_{h}=\phi(\pmb{Z}^{(i)}_{h-2}, \pmb{Z}^{(i)}_{h-1};\{\underbrace{\pmb{W}^{(i)_{deep}}_h}_{stride=1}, \pmb{\beta}^{(i)_{deep}}_h\}), i=1,2,\dots, u, \label{deep}
\end{align}
\begin{align}
\pmb{Z}^{(i)}_{k+1}=\phi(\pmb{Z}^{(i)}_{k-1}, \pmb{Z}^{(i)}_{k};\{\underbrace{\pmb{W}^{(i)_{broad}}_{k+1}}_{stride=2}, \pmb{\beta}^{(i)_{broad}}_{k+1}\}), i=1,2,\dots, u, \label{broad}
\end{align}
where $\{\pmb{W}^{(i)_{deep}}_h, \pmb{\beta}^{(i)_{deep}}_h\}$ and $\{\pmb{W}^{(i)_{broad}}_{k+1}, \pmb{\beta}^{(i)_{broad}}_{k+1}\}$ are the weight, bias matrices of deep Cells and broad Cell in convolution block $i$, respectively. Moreover, $\phi(\cdot)$ is a set of transformations (e.g. depthwise-separable convolution \cite{chollet2017xception}, pooling, skip connection) by the deep Cells and broad Cell. In other words, each Cell in the convolution block uses the outputs of its previous two Cells as the inputs for combining various features. However, there is a doubt in (\ref{deep}) that $\pmb{Z}^{(i)}_{-1}$ and $\pmb{Z}^{(i)}_{0}$ are not defined. A complementary expression is given as
\begin{align}
\{\pmb{Z}^{(i)}_{-1}, \ \pmb{Z}^{(i)}_{0}\}=\{\pmb{Z}^{(i-1)}_{k}, \ \pmb{Z}^{(i-1)}_{k+1}\}, i=2,3,\dots,u. \label{complementary}
\end{align}
Moreover, a convolution with $3\times3$ kernel size is inserted as a stem of BCNN to provide the input information for the first and second convolution Cell. As a result, the output of $3\times3$ convolution can be represented as $\pmb{Z}^{(1)}_\xi$, where $\xi\leq0$.

\begin{figure}[!ht]
\centering
\includegraphics[width=6.5cm]{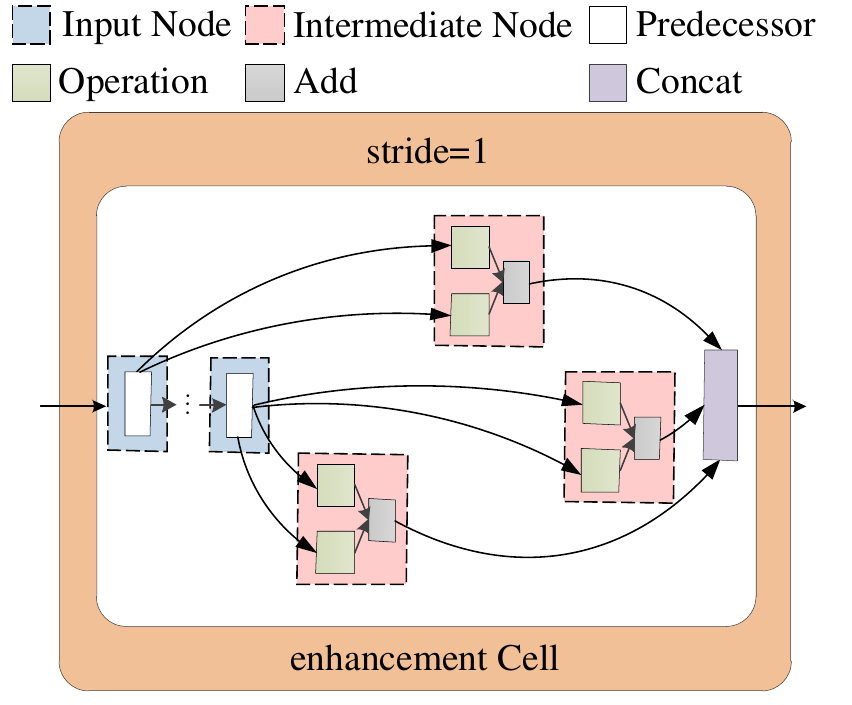}
\caption{The structure of enhancement block.}
\label{enhancement_block}
\end{figure}

\paragraph{Enhancement Block} The structure of enhancement block is shown in Fig. \ref{enhancement_block}. In each enhancement block, there is a single enhancement Cell with one stride and different topology from those convolution Cells. For enhancement block $En\_j \; (j=1,2,\dots,v)$, its enhancement feature representations $\pmb{H}^{(j)}$ can be mathematically expressed as
\begin{align}
\pmb{H}^{(j)}=\varphi(\delta(\pmb{Z}^{(1)}_{0}, \pmb{Z}^{(1)}_{k+1}, \cdots, \pmb{Z}^{(u-1)}_{k+1}), \pmb{Z}^{(u)}_{k+1};\{\underbrace{\pmb{W}^{(j)}}_{stride=1}, \pmb{\beta}^{(j)}\}), \label{en_block}
\end{align}
where $\pmb{W}^{(j)}$ and $\pmb{\beta}^{(j)}$ are the weight and bias matrices of enhancement Cell in $En\_j$, respectively. Moreover, $\varphi(\cdot)$ is a set of transformations by the enhancement Cell, and $\delta(\cdot)$ is a function combination of $1\times1$ convolution and concatenating.

\paragraph{Multi-scale Feature Fusion} For achieving promised performance of BCNN, all outputs of each convolution and enhancement block (with various feature scales) are connected directly with the global average pooling (GAP) layer to yield more reasonable and comprehensive representations. Here, the output of the last deep Cell in each convolution block is connected for feeding multi-scale features into the GAP layer, so that the final output of GAP layer can be expressed as
\begin{align}
\pmb{O}=\psi(\pmb{Z}^{(1)}_{k}, \pmb{Z}^{(2)}_{k}, \cdots, \pmb{Z}^{(u)}_{k}, \pmb{H}^{(1)}, \pmb{H}^{(2)}, \cdots, \pmb{H}^{(v)}), \label{gap}
\end{align}
where $\psi(\cdot)$ is a function combination of $1\times1$ convolution, concatenating and global average pooling.

\begin{figure}[!ht]
\centering
\includegraphics[width=8cm]{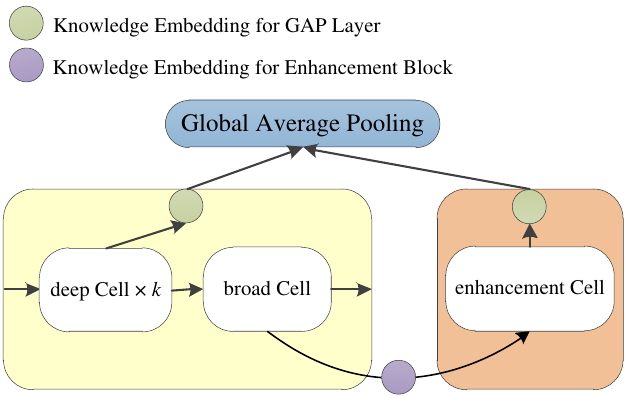}
\caption{Prior knowledge embedding.}
\label{pke}
\end{figure}

\paragraph{Prior Knowledge Embedding} A priori knowledge whose position is shown in Fig. \ref{pke}, is incorporated into the proposed BCNN. Depending on substantial experiments, we find that those low-pixels feature maps are more important than those feature maps with high resolutions for achieving high performance. In other words, for designing BCNN with novel performance, more deep and broad feature maps of $Conv\_r$ instead of $Conv\_s$ should be fed into the enhancement block, where $r>s$ and $0<s,r \leq u$. In order to insert the above priori knowledge into BCNN, a group of convolutions with $1\times1$ kernel size are employed in each connection between the convolution block (except the last one) and enhancement block. These $1\times1$ convolutions accept those feature representations from the final deep Cell in each convolution block, and output a group of feature maps with different importance. Moreover, the importance is represented by the number of output channels which the larger is the more important it is. Furthermore, these $1\times1$ convolutions employ different strides for concatenating all input feature maps with same size. As a result, the output of $\delta(\pmb{Z}^{(1)}_{0}, \pmb{Z}^{(1)}_{k+1}, \cdots, \pmb{Z}^{(u-1)}_{k+1})$ in \eqref{en_block} is obtained by concatenating those outputs of $1\times1$ convolutions whose inputs are $\pmb{Z}^{(1)}_{0}, \pmb{Z}^{(1)}_{k+1}, \cdots, \pmb{Z}^{(u-1)}_{k+1}$.

In order to discovering the broad scalable architecture with satisfactory performance and extreme fast efficiency, an efficient optimization strategy is necessary. In the following, we will introduce the efficient optimization strategy adopted in BNAS.


\begin{figure}[!t]
\centering
\includegraphics[width=8cm]{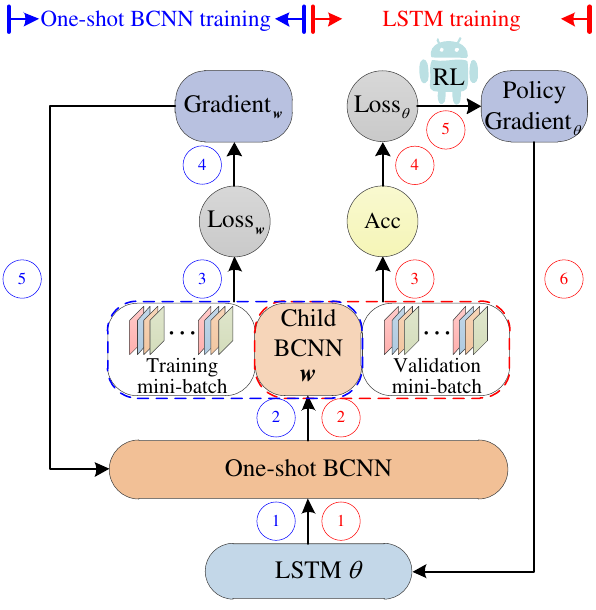}
\caption{Overview of efficient optimization strategy of BNAS. There are two interleaving phases, one-shot BCNN and LSTM training, for architecture search. In the first phase, one-shot BCNN contained all possible child BCNN is training with parameters sharing mechanism. In the second phase, policy gradient-based reinforcement learning algorithm is employed to update the LSTM for discovering better architecture.}
\label{optimization}
\end{figure}

\subsection{Efficient Optimization Strategy}
\label{Optimization Strategy}

In BNAS, the combination of parameter sharing presented in ENAS \cite{pham2018efficient} and reinforcement learning is adopted for finding broad scalable architecture with satisfactory performance. Here, an Long Short-Term Memory (LSTM) \cite{hochreiter1997long} controller with parameter $\theta$ is trained in a loop with two phases: one-shot BCNN training and LSTM training whose details can be found in Fig. \ref{optimization}. Moreover, the one-shot BCNN defines all possible broad scalable architectures under a set of pre-defined candidate operations.

In the first phase, the one-shot BCNN is trained in a single epoch, where each step sample a child BCNN from the one-shot BCNN by the LSTM:
\begin{enumerate}
  \item LSTM first samples two types of Cells, convolution Cell and enhancement Cell, with a list of tokens $a_{1:T}$ according a sampling policy $\pi(\cdot)$.
  \item A child BCNN $m$ can be determined according to the sampled convolution Cell and enhancement Cell, and its weights $\pmb{w}$ is inherited from the one-shot BCNN.
  \item A training mini-batch data is fed into $m$ (a sub-network of one-shot BCNN) to obtain a loss.
  \item The above loss is used to compute the gradient with respect to the one-shot BCNN.
  \item Update the sample part of the one-shot BCNN.
\end{enumerate}

In the second phase, the LSTM is trained by reinforcement learning in fixed steps:
\begin{enumerate}
  \item LSTM first samples two types of Cells, convolution Cell and enhancement Cell, with a list of tokens $a_{1:T}$ according a sampling policy $\pi(\cdot)$.
  \item A child BCNN $m$ can be determined according to the sampled convolution Cell and enhancement Cell, and its weights $\pmb{w}$ is inherited from the one-shot BCNN.
  \item A validation mini-batch data is fed into $m$ to obtain an accuracy $R(m;\pmb{w})$.
  \item According to $R(m;\pmb{w})$, a loss with respect to $\theta$ can be obtained on the desired task.
  \item Reinforcement learning is employed to calculate the policy gradient respect to $\theta$.
  \item Using the above policy gradient to update the LSTM also the sampling policy $\pi(\cdot)$ for discovering various Cells with better performance.
\end{enumerate}

Here, BNAS demands the LSTM controller to maximize the expected reward $J(\theta)$, where
\begin{align}
J(\theta)=\mathbb{E}_{\pi(a_{1:T}; \theta)}[R(m;\pmb{w})]. \label{expected reward}
\end{align}
Moreover, a gradient policy algorithm, REINFORCE \cite{williams1992simple} is applied to compute the policy gradient $\nabla_{\theta}J(\theta)$, where
\begin{align}
\nabla_{\theta}J(\theta)=\sum_{t=1}^{T}\mathbb{E}_{\pi(a_{1:T}; \theta)}[\nabla_{\theta}log\pi(a_t|a_{(t-1):1};\theta)R(m;\pmb{w})].
\label{policy gradient}
\end{align}

After many iterations of this loop are repeated, novel Cells with satisfactory performance can be found.

\subsection{Comparison of BCNN and BLS}
\label{problem}

Inspired by the topologies of BLS and its variants CCFBLS and CCEBLS, we propose BCNN and several variants with different topologies (see Section \ref{variants}). Therefore, the proposed BCNN is of similar topology with BLS rather than other characteristics.

\subsubsection{Element} The primary component of BCNN is Cell with powerful feature extraction ability that consists of convolution, pooling and skip connection, instead of neuron in BLS.

\subsubsection{Connection} For promising performance for image classification, prior knowledge embedding is inserted in the connections between convolution Cell and enhancement Cell, convolution Cell and GAP layer, and enhancement Cell and GAP layer.

\subsubsection{Topology} BCNN and its variants are proposed with the motivation of traditional BLS ant its variants with different topologies. However, they are not identical in terms of topology. In BLS, training data is fed into each group of mapped feature nodes as input. BCNN only take training data in the first convolution block as its input. Moreover, other convolution blocks take outputs of previous two blocks as input, i.e. BCNN is not only broad but also deep.

\subsubsection{Design} Different from human designed BLS, the proposed BCNN is designed in an automatic way. In other word, BNAS can design appropriate BCNN for specific task with satisfactory performance.

\subsubsection{Optimization} Incremental learning algorithms are employed to optimize the weights of BLS. Differently, we adopt gradient descent-based algorithm to update the weights of BCNN iteratively.

\section{Universal Approximation Property of Broad Convolutional Neural Network}
\label{approximation}
As a new paradigm of deep neural networks, we also analyze the universal approximation ability of BCNN, and prove a theorem to verify the effectiveness of the proposed architecture.

As described in Section \ref{bcnn}, the input of GAP layer comes from all convolution and enhancement blocks with various significance, i.e. number of channels. After the processing of GAP layer, each channel can be treated as the output of a neuron with respect to input data $\pmb{x}$. Here, we assume that the number of channels of $\pmb{Z}^{(1)}_{k}$ from the first convolution block is $c$. As a result, the number of channels of the outputs of convolution blocks $\pmb{Z}^{(i)}_{k}(i=2,\dots, u)$ and enhancement blocks $\pmb{H}^{(j)}(j=1, \dots, v)$ are $c \times 2^{(i-1)}$ and $c \times 2^{j}$, respectively. Accordingly, $u$ convolution and $v$ enhancement blocks provide $U=c \times (2^u-1)$ and $V=c \times v \times 2^u$ channels (i.e. the output of neurons) for fully connected layer, respectively.

For standard hypercube $\textbf{I}^d = [0;1]^d$ of $\mathbb{R}^d$, any continuous function $f \in C(\textbf{I}^d)$ and activation function $\sigma$, the BCNN can be equivalently expressed as
\begin{align}
\begin{aligned}
f_{\pmb{p}_{u,v}}=&\sum_{i=1}^{U}w_i\sigma(\pmb{x}; \{\phi,\psi, \pmb{W}^{conv}_{i}, \pmb{\beta}^{conv}_{i}\})\\
&+\sum_{j=1}^{V}w_{U+j}\sigma(\pmb{x}; \{\delta,\varphi,\psi,\pmb{W}^{en}_{j}, \pmb{\beta}^{en}_{j}\}),
\end{aligned}
\end{align}
where, $\pmb{p}_{u,v}=(u,v,c,w_1,\dots,w_{U+V},\pmb{W}^{conv}, \pmb{\beta}^{conv},\pmb{W}^{en},$ $\pmb{\beta}^{en})$ is the set of overall parameters for the proposed BCNN, $\{\pmb{W}^{conv}, \pmb{\beta}^{conv}\}$ and $\{\pmb{W}^{en}, \pmb{\beta}^{en}\}$ are the corresponding parameters of convolution and enhancement blocks, respectively. Moreover, we employ $\pmb{\xi}_{u,v}=(w_1,\dots,w_{U+V},\pmb{W}^{conv}, \pmb{\beta}^{conv},$ $\pmb{W}^{en}, \pmb{\beta}^{en})$ to represent the randomly generated variables that are defined on the probability measure $\zeta_{u,v}$. For compact set $\Omega$ of $\textbf{I}^d$, the distance between the continuous function $f$ and $f_{\pmb{p}_{u,v}}$ can be computed as
\begin{align}
\chi_{\Omega}(f,f_{\pmb{p}_{u,v}})=\sqrt{\mathbb{E}\left[ \int_{\Omega}(f(\pmb{x})-f_{\pmb{p}_{u,v}}(\pmb{x}))^2d\pmb{x} \right]}.
\end{align}
Based on the above hypotheses, we present our results as below.

\emph{Theorem 1}: For any continuous function $f \in C(\textbf{I}^d)$ and any compact set $\Omega$ of $\textbf{I}^d$, BCNN with nonconstant bounded feature mappings $\phi,\delta,\varphi,\psi$, and absolutely integrable activation function $\sigma$ (defined on $\textbf{I}^d$ so that $\int_{\mathbb{R}_d}\sigma^2(\pmb{x})d\pmb{x}< \infty$), has a sequence of $\{f_{\pmb{p}_{u,v}}\}$ with regard to a corresponding sequence of probability measures $\zeta_{u,v}$, such that
\begin{align}
\mathop{\rm{lim}}\limits_{u,v\rightarrow \infty}\chi_{\Omega}(f,f_{\pmb{p}_{u,v}})=0.
\end{align}
Moreover, the trainable set of parameters $\pmb{\xi}_{u,v}$ is generated by the distributions of $\zeta_{u,v}$.

\emph{Proof}: Give several definitions that the continuous function $f \in C(\textbf{I}^d)$, the approximation function $f_{\pmb{p}_{u,v}}$ of BCNN, the weight matrix between fully connected layer and the output of GAP layer with respect to those feature maps from convolution blocks  $\pmb{w}_c=[w_{c1},\dots,w_{cU}]$, the weight matrix between fully connected layer and the output of GAP layer with respect to those feature maps from enhancement blocks $\pmb{w}_e=[w_{e1},\dots,w_{eV}]$.

For a BCNN with $u$ (any integer) convolution blocks, we define
\begin{align}
f_{\pmb{w}_{c}}=&\sum_{i=1}^{U}w_{ci}\sigma(\pmb{x}; \{\phi,\psi, \pmb{W}^{conv}_{i}, \pmb{\beta}^{conv}_{i}\}),
\end{align}
where trainable parameters are sampled from the predefined $\zeta_{u,v}$. Since the feature mappings $\phi,\psi$ are nonconstant bounded, the resident function of input $\pmb{x}$
\begin{align}
f_{r_u}(\pmb{x}) = f(\pmb{x})-f_{\pmb{w}_{c}}(\pmb{x})
\end{align}
defined on $\textbf{I}^d$ is bounded and integrable. According to the fact provided in \cite{rudin2006real}, for $\forall \varepsilon>0$, we always can find a function $f_{b_u} \in C(\textbf{I}^d)$ that satisfies the following expression as
\begin{align}
\chi_{\Omega}(f_{b_u},f_{r_u})<\frac{\varepsilon}{2}.
\label{buru}
\end{align}
Moreover, detailed discussions can be found in \cite{chen2018universal}.

In order to approximate $f_{b_u}$, we define
\begin{align}
f_{\pmb{w}_{e}}=&\sum_{j=1}^{V}w_{ej}\underbrace{\sigma(\pmb{x}; \{\delta,\varphi,\psi, \pmb{W}^{en}_{j}, \pmb{\beta}^{en}_{j}\})}_{\vartheta_j},
\label{fwe}
\end{align}
where trainable parameters are sampled from the predefined $\zeta_{u,v}$. Similarly, the composition function $\vartheta_{j} (j=1, \dots, V)$ in \eqref{fwe} is absolutely integrable due to nonconstant bounded feature mappings $\delta,\varphi,\psi$. According to \emph{Theorem 1} in \cite{igelnik1995stochastic}, for $\forall \varepsilon>0$, we always can find a sequence of $f_{\pmb{w}_{e}}$ that satisfies the following expression as
\begin{align}
\chi_{\Omega}(f_{b_u},f_{\pmb{w}_{e}})<\frac{\varepsilon}{2}.
\end{align}

At last, the distance between the continuous function $f$ and the approximation function $f_{\pmb{p}_{u,v}}$ of BCNN can be computed as
\begin{align}
\begin{aligned}
\chi_{\textbf{I}^d}(f,f_{\pmb{p}_{u,v}})&=\sqrt{\mathbb{E}\left[ \int_{\Omega}(f(\pmb{x})-f_{\pmb{p}_{u,v}}(\pmb{x}))^2d\pmb{x} \right]}\\
&= \sqrt{\mathbb{E}\left[ \int_{\Omega}\left((f(\pmb{x})-f_{\pmb{w}_{c}}(\pmb{x}))-f_{\pmb{w}_{e}}\right)^2d\pmb{x} \right]}\\
&= \sqrt{\mathbb{E}\left[ \int_{\Omega}\left(f_{r_u}-f_{\pmb{w}_{e}}\right)^2d\pmb{x} \right]}\\
&= \chi_{\Omega}(f_{r_u},f_{\pmb{w}_{e}})\\
&\leq \chi_{\Omega}(f_{b_u},f_{r_{u}}) + \chi_{\Omega}(f_{b_u},f_{\pmb{w}_{e}}) \\
&< \frac{\varepsilon}{2} + \frac{\varepsilon}{2} \\
&=\varepsilon\\
\end{aligned}
\end{align}
Therefore, we can conclude that
\begin{align}
\mathop{\rm{lim}}\limits_{u,v\rightarrow \infty}\chi_{\Omega}(f,f_{\pmb{p}_{u,v}})=0.
\end{align}


\begin{figure*}[!t]
\centering
\includegraphics[width=16.3cm]{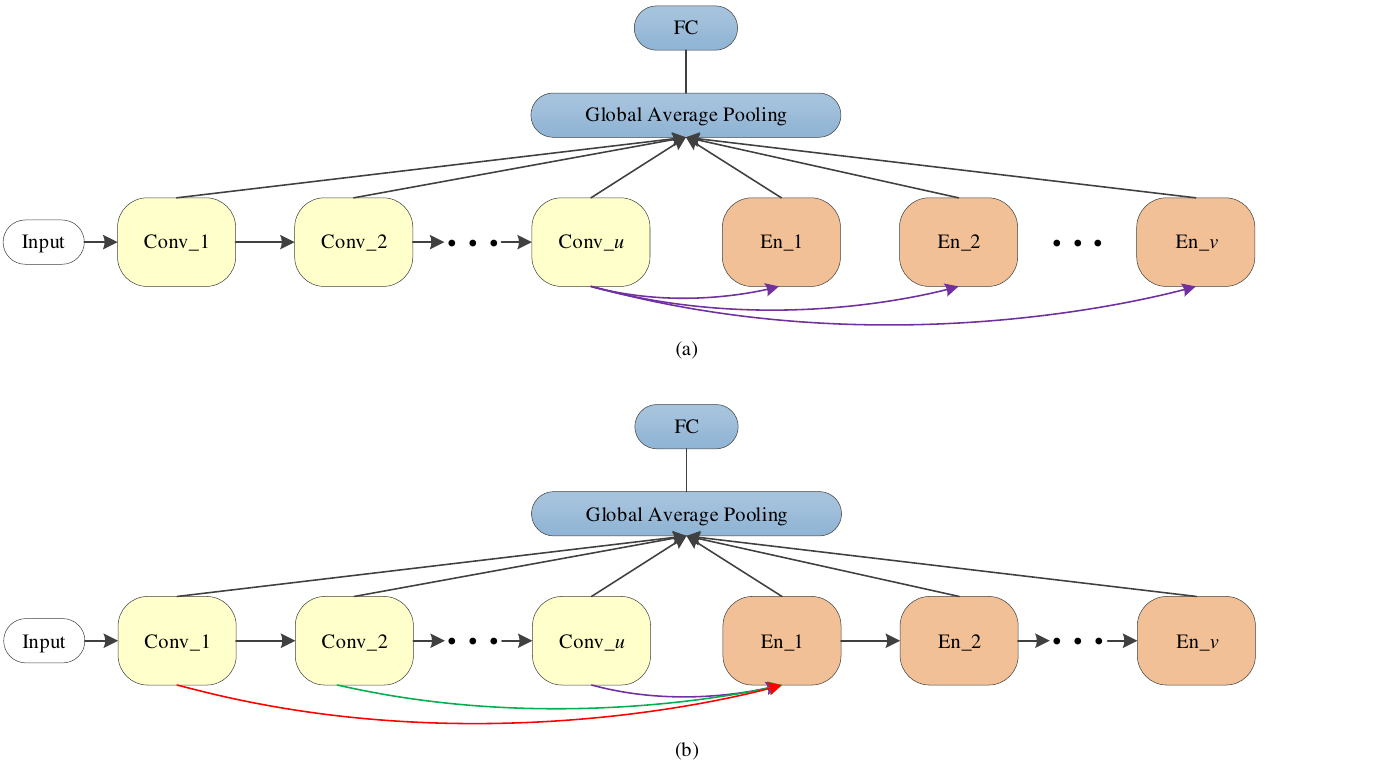}
\caption{The topologies of broad scalable architectures employed in BNAS-CCLE and BNAS-CCE. (a) The topology of broad scalable architecture employed in BNAS-CCLE. (b) The topology of broad scalable architecture employed in BNAS-CCE.}
\label{BCNN-CCLE-CCE}
\end{figure*}

\section{Variants of Broad Neural Architecture Search}
\label{variants}
BLS is a flexible paradigm to be modified under different constraints. Moreover, several variants have been proposed in \cite{chen2018universal}, e.g., CCFBLS and CFEBLS. Motivated by the above two variants of BLS, we also propose two developed versions for BNAS who employ various topologies of BCNN: 1) Cascade of Convolution blocks with its Last block connected to the Enhancement blocks Broad Neural Architecture Search (BNAS-CCLE) and 2) Cascade of Convolution blocks and Enhancement blocks Broad Neural Architecture Search (BNAS-CCE). In this section, we only introduce the difference between BNAS and its two variants, i.e. the broad scalable architectures used in BNAS-CCLE and BNAS-CCE which are shown in Fig. \ref{BCNN-CCLE-CCE}.

\subsection{BNAS-CCLE}

The broad scalable architecture of BNAS-CCLE is a developed CCFBLS \cite{chen2018universal} which is not only broad but also deep. In the broad scalable architecture of BNAS-CCLE, each enhancement block only treats the output of last convolution block as input. In other words, we assume that the output of last convolution block takes up all importance of each enhancement block.


Related expressions with regard to convolution blocks can be found in (\ref{deep}) to (\ref{complementary}). For enhancement block $En\_j$, its enhancement feature representations $\pmb{H}^{(j)}$ can be defined as
\begin{align}
\pmb{H}^{(j)}=\varphi(\pmb{Z}^{(u)}_{k}, \pmb{Z}^{(u)}_{k+1};\{\pmb{W}^{(j)}, \pmb{\beta}^{(j)}\}), j=1,2,\dots,v \label{en_block_2}
\end{align}
where $\pmb{W}^{(j)}$ and $\pmb{\beta}^{(j)}$ are the weight and bias matrices of enhancement Cell in enhancement block $j$, respectively. Similarly, $\varphi(\cdot)$ is a set of transformations by the enhancement Cell.

For performance improvement, we also incorporate aforementioned priori knowledge used in BCNN into the connections between convolution blocks and GAP layer. On one hand, the importance of each convolution block is twice as much as its previous one for GAP layer. For instance, the GAP layer accepts $a$ and $2a$ channels from $Conv\_x$ and $Conv\_y \; (y=x+1)$, respectively. On the other hand, the importance of each enhancement block is equal. For instance, the GAP layer accepts $b$ channels from both $En\_x$ and $En\_y \; (1 \leq x,y \leq v)$.

\subsection{BNAS-CCE}

The broad scalable architecture of BNAS-CCE is a developed CCEBLS \cite{chen2018universal} that employs deeper topology than other variants of BLS. There are two differences between the broad scalable architectures of BNAS and BNAS-CCLE: 1) The output of each convolution block is fed into the first enhancement block as its inputs. 2) Enhancement blocks are stacked one after another rather than in a parallel way.

\begin{table*}[!t]
\centering
\begin{threeparttable}[h]
\caption{Hyper-parameters determination for BNAS and BNAS-CCE $\dag$}
\label{hyper-parameters-bnas}
\begin{tabular}{lccccccccc}
\hline

\multirow{2}{*}{\textbf{Approach}} & \multirow{2}{*}{$\pmb{v}_{s}$} & \multirow{2}{*}{$\pmb{v}_{d}$} & \multirow{2}{*}{$\pmb{k}_{d}$} & \multicolumn{5}{c}{\textbf{Accuracy (\%)}} & \multirow{2}{*} {\textbf{Mean $\pm$ Var (\%)}} \\
\cline{5-9}
 &  &   &   & \textbf{Arch-1} & \textbf{Arch-2} & \textbf{Arch-3} & \textbf{Arch-4} & \textbf{Arch-5} & \\
\hline
\hline
\multirow{18}{*}{BNAS} & \multirow{9}{*}{1} & \multirow{3}{*}{1} & 0 & 94.56 & 95.50 & 95.51 & 95.64 & 95.78 & 95.40 $\pm$ 0.19\\
\cline{4-10}
 &  &  & 1 & 96.42 & 96.63 & 96.56 & 96.21 & 96.64 &96.49 $\pm$ 0.03\\
\cline{4-10}
 &  &  & 2 & 96.32 & 96.78 & 96.51 & 96.58 & 96.32 &96.50 $\pm$ 0.03\\
\cline{3-10}
 &  & \multirow{3}{*}{2} & 0 & 94.34 & 95.53 & 95.60 & 96.02 & 95.80 & 95.46 $\pm$ 0.34\\
\cline{4-10}
 & &  & 1 & 96.40 & 96.56 & 96.36 & 96.48 & 96.37 & 96.43 $\pm$ 0.01\\
\cline{4-10}
 &  &  & 2 & 95.85 & 96.65 & 96.51 & 96.41 & 96.58 & 96.40 $\pm$ 0.08\\
\cline{3-10}
 &  & \multirow{3}{*}{3} & 0 & 93.95 & 95.75 & 95.57 & 95.37 & 94.97 &95.12 $\pm$ 0.41\\
\cline{4-10}
 &  &  & 1 & 96.49 & 96.59 & 96.53 & 96.24 & 96.50 & 96.47 $\pm$ 0.01\\
\cline{4-10}
 &  &  & 2 & 96.57 & 96.79 & 96.01 & 96.56 & 96.64 & 96.51 $\pm$ 0.07\\
\cline{2-10}
 & \multirow{9}{*}{\textbf{2}} & \multirow{3}{*}{\textbf{1}} & 0 & 95.20 & 95.65 & 94.74 & 95.86 & 95.11 & 95.31 $\pm$ 0.16\\
\cline{4-10}
 &  &  & \textbf{1} & 96.86 & 96.82 & 96.28 & 96.77 & 96.79 & \textbf{96.70 $\pm$ 0.05}\\
\cline{4-10}
 &  &  & 2 & 96.83 & 96.36 & 96.43 & 96.76 & 96.72 &96.62 $\pm$ 0.04\\
\cline{3-10}
 &  & \multirow{3}{*}{2} & 0 & 94.72 & 93.89 & 93.12 & 95.56 & 95.72 & 94.60 $\pm$ 0.98\\
\cline{4-10}
 &  &  & 1 & 96.79 & 96.46 & 96.45 & 96.56 & 96.49 &96.55 $\pm$ 0.02\\
\cline{4-10}
 &  &  & 2 & 96.73 & 96.37 & 96.48 & 96.72 & 96.74 & 96.61 $\pm$ 0.02\\
\cline{3-10}
 &  & \multirow{3}{*}{3} & 0 & 95.44 & 95.52 & 95.28 & 95.55 & 94.85 & 95.33 $\pm$ 0.07\\
\cline{4-10}
 &  &  & 1 & 96.70 & 96.72 & 96.47 & 96.55 & 96.66 & 96.62 $\pm$ 0.01\\
\cline{4-10}
 &  &  & 2 & 96.76 & 96.01 & 96.63 & 96.50 & 96.82 &96.54 $\pm$ 0.08\\
\hline
\hline

\multirow{18}{*}{BNAS-CCE} & \multirow{9}{*}{1} & \multirow{3}{*}{1} & 0 & 95.24 & 95.34 & 94.11 & 93.76 & 94.74 & 94.64 $\pm$ 0.38\\
\cline{4-10}
 & &  & 1 & 96.08 & 95.91 & 96.07 & 96.17 & 95.55 &95.96 $\pm$ 0.05\\
\cline{4-10}
 & &  & 2 & 96.40 & 95.79 & 95.96 & 96.42 & 96.68 &96.25 $\pm$ 0.11\\
\cline{3-10}
 & & \multirow{3}{*}{2} & 0 & 94.89 & 95.41 & 94.91 & 94.78 & 95.05 & 95.01 $\pm$ 0.05\\
\cline{4-10}
 & &  & 1 & 95.82 & 95.89 & 96.13 & 96.00 & 95.33 & 95.83 $\pm$ 0.08\\
\cline{4-10}
 & &  & 2 & 96.38 & 96.55 & 95.94 & 96.15 & 95.83 & 96.17 $\pm$ 0.07 \\
\cline{3-10}
 & & \multirow{3}{*}{3} & 0 & 95.73 & 95.42 & 95.43 & 95.26 & 95.53 &95.47 $\pm$ 0.02\\
\cline{4-10}
 & &  & 1 & 96.30 & 95.93 & 96.46 & 96.25 & 95.92 &96.17 $\pm$ 0.05\\
\cline{4-10}
 & &  & 2 & 96.42 & 95.74 & 96.07 & 95.94 & 96.01 & 96.04 $\pm$ 0.05\\
\cline{2-10}
 & \multirow{9}{*}{\textbf{2}} & \multirow{3}{*}{1} & 0 & 95.51 & 95.51 & 95.03 & 95.85 & 95.65 & 95.51 $\pm$ 0.07\\
\cline{4-10}
 & &  & 1 & 96.22 & 96.17 & 96.67 & 96.53 & 96.40 &96.40 $\pm$ 0.04\\
\cline{4-10}
 & &  & 2 & 95.62 & 96.73 & 96.40 & 96.64 & 96.68 &96.41 $\pm$ 0.17\\
\cline{3-10}
 & & \multirow{3}{*}{\textbf{2}} & 0 & 95.06 & 95.78 & 94.74 & 95.38 & 95.61 &95.31 $\pm$ 0.14\\
\cline{4-10}
 & &  & 1 & 96.32 & 96.11 & 95.91 & 96.39 & 96.42 & 96.23 $\pm$ 0.04 \\
\cline{4-10}
 & &  & \textbf{2} & 96.71 & 96.56 & 96.13 & 96.63 & 96.77 &\textbf{96.56 $\pm$ 0.05}\\
\cline{3-10}
 & & \multirow{3}{*}{3} & 0 & 95.39 & 95.53 & 95.53 & 95.31 & 95.87 &95.53 $\pm$ 0.04\\
\cline{4-10}
 & &  & 1 & 95.70 & 96.07 & 96.25 & 96.54 & 96.10 &96.13 $\pm$ 0.07\\
\cline{4-10}
 & &  & 2 & 96.44 & 96.63 & 96.48 & 96.41 & 96.56 &96.50 $\pm$ 0.01\\
\hline
\end{tabular}
\footnotesize
\begin{tablenotes}
\item[$\dag$] We only show the experimental results with respect to BNAS and BNAS-CCE, due to similar performance of BNAS and BNAS-CCLE for hyper-parameters determination.
\end{tablenotes}
\end{threeparttable}
\end{table*}

Similarly, related expressions with regard to convolution blocks can be represented by (\ref{deep}) to (\ref{complementary}). Furthermore, enhancement feature representations $\pmb{H}^{(j)} \; (j=1,2,\dots,v)$ can be divided into three cases:

\begin{align}
\left\{
\begin{aligned}
&\varphi(\delta(\pmb{Z}^{(1)}_{k+1}, \cdots, \pmb{Z}^{(u-1)}_{k+1}), \pmb{Z}^{(u)}_{k+1}); \{\pmb{W}^{(j)}, \pmb{\beta}^{(j)}\}), &if \; j = 1\\
&\varphi(\pmb{Z}^{(u)}_{k+1}, \pmb{H}^{(1)};\{\pmb{W}^{(j)}, \pmb{\beta}^{(j)}\}), &if \; j = 2\\
&\varphi(\pmb{H}^{(j-2)}, \pmb{H}^{(j-1)};\{\pmb{W}^{(j)}, \pmb{\beta}^{(j)}\}), &else\\
\end{aligned}
\right.
\label{en_block_3}
\end{align}
where $\pmb{W}^{(j)}$ and $\pmb{\beta}^{(j)}$ are the weight and bias matrices of enhancement Cell in enhancement block $j$, respectively. Similarly, $\varphi(\cdot)$ is a set of transformations by the enhancement Cell. And $\delta(\cdot)$ is aforementioned function combination of $1\times1$ convolution and concatenating.

Similar with BNAS and BNAS-CCLE, for the broad scalable architecture of BNAS-CCE, $1\times1$ convolution- based priori knowledge is inserted between convolution blocks and the first enhancement block, also is inserted into the connections of convolution blocks and GAP layer. Different from BNAS and BNAS-CCLE, the importance of each enhancement block in broad scalable architecture of BNAS-CCE is not equal for GAP layer. For instance, the GAP layer accepts $c$ channels from both $En\_x$ and $En\_y \; (1 \leq x,y < v, y=x+1)$, and all channels from $En\_v$.


\section{Experiments and Analysis}
\label{experiments and analysis}

\begin{table*}[!t]
\centering
\begin{threeparttable}[tbq]
\caption{Comparison of the proposed BNAS with other NAS approaches on CIFAR-10 for small-size model.}
\label{cifar10_s}
\begin{tabular}{lccccc}
\hline
\multirow{2}{*}{\textbf{Architecture}} & \textbf{Error}& \textbf{Params}& \textbf{Search Cost} &\multirow{2}{*}{\textbf{Search Method}}&\multirow{2}{*}{\textbf{Topology}}\\
                                       & \textbf{(\%)} & \textbf{(M)}   & \textbf{(GPU Days)}  &\\
\hline
LEMONADE + cutout \cite{elsken2018efficient}    & 4.57 & 0.5  & 80  & evolution & deep\\
DPP-Net + cutout \cite{dong2018dpp}             & 4.62 & 0.5  & 4.00& evolution  & deep \\
\hline
BNAS + cutout (ours)                      & 3.83 & 0.5 & 0.20 & RL & broad\\
BNAS-CCLE + cutout (ours)                 & 3.63 & 0.5 & 0.20 & RL & broad\\
BNAS-CCE + cutout (ours)                  & \textbf{3.58} & 0.6 & \textbf{0.19} & RL & broad\\
\hline
\end{tabular}
\end{threeparttable}
\end{table*}

\begin{table*}[!t]
\centering
\begin{threeparttable}[tbq]
\caption{Comparison of the proposed BNAS with other NAS approaches on CIFAR-10 for medium-size model.}
\label{cifar10_m}
\begin{tabular}{lccccc}
\hline
\multirow{2}{*}{\textbf{Architecture}} & \textbf{Error}& \textbf{Params}& \textbf{Search Cost} &\multirow{2}{*}{\textbf{Search Method}}&\multirow{2}{*}{\textbf{Topology}}\\
                                       & \textbf{(\%)} & \textbf{(M)}   & \textbf{(GPU Days)}  &\\
\hline
LEMONADE + cutout \cite{elsken2018efficient}    & 3.69 & 1.1  & 80  & evolution & deep\\
DPP-Net + cutout \cite{dong2018dpp}             & 4.78 & 1.0  & 4.00& evolution & deep\\
\hline
BNAS + cutout (ours)                      & 3.46 & 1.1  & 0.20 & RL & broad\\
BNAS-CCLE + cutout (ours)                  & 3.40 & 1.1  & 0.20 & RL & broad\\
BNAS-CCE + cutout (ours)                  & \textbf{3.24} & 1.0  & \textbf{0.19} & RL & broad\\
\hline
\end{tabular}
\end{threeparttable}
\end{table*}

In this section, we perform a set of experiments to examine several novel properties of BNAS and its two variants. First of all, a large number of experiments are performed by BNAS and its two variants for hyper-parameters (e.g., the number of enhancement block in two phases, the number of deep Cell in architecture estimation phase) determination. Next, BNAS and its two variants are applied to discover novel architecture on CIFAR-10 for testing the search efficiency and high performance of the learned architecture. Subsequently, the learned architecture with best performance is chosen to solve large scale image classification task on ImageNet. The experiment on ImageNet not only evaluates the transferability of the discovered architectures of BNAS and its two variants, but also examines the powerful multi-scale features extraction capacity of the proposed broad scalable architectures. Finally, the qualitative and quantitative analysis are given for the experimental results on CIFAR-10 and ImageNet.

\begin{table*}[!t]
\centering
\begin{threeparttable}[tbq]
\caption{Comparison of the proposed BNAS with other NAS approaches on CIFAR-10 for large-size model.}
\label{cifar10_l}
\begin{tabular}{lccccc}
\hline
\multirow{2}{*}{\textbf{Architecture}} & \textbf{Error}& \textbf{Params}& \textbf{Search Cost} &\multirow{2}{*}{\textbf{Search Method}}&\multirow{2}{*}{\textbf{Topology}}\\
                                       & \textbf{(\%)} & \textbf{(M)}   & \textbf{(GPU Days)}  &\\
\hline
AmoebaNet-A + cutout \cite{real2018regularized} & 3.34 $\pm$ 0.06 & 3.2  & 3150 & evolution & deep\\
AmoebaNet-B + cutout \cite{real2018regularized} & 2.55 $\pm$ 0.05 & 2.8  & 3150 & evolution & deep\\
Hierarchical Evo \cite{liu2017hierarchical}     & 3.75 $\pm$ 0.12 & 15.7  & 300 & evolution & deep\\
LEMONADE + cutout \cite{elsken2018efficient}    & 3.05 & 4.7  & 80  & evolution & deep\\
DARTS(second order) + cutout \cite{liu2018darts}& 2.83 $\pm$ 0.06 & 3.3  & 4.00& gradient-based & deep\\
DARTS(first order) + cutout \cite{liu2018darts} & 3.00 & 2.9  & 1.50 & gradient-based & deep \\
P-DARTS + cutout \cite{chen2019progressive}  & \textbf{2.50} & 3.4  & 0.30 & gradient-based & deep \\
PC-DARTS + cutout \cite{xu2019pc} & 2.57 $\pm$ 0.07 & 3.6  & \textbf{0.10} & gradient-based & deep \\
NASNet-A + cutout \cite{zoph2018learning}       & 2.65 & 3.3  & 1800 & RL & deep\\
NASNet-B + cutout \cite{zoph2018learning}       & 3.73 & \textbf{2.6}  & 1800 & RL & deep\\
MANAS + cutout \cite{carlucci2019manas}         & 2.63 & 3.4  & 2.80 & RL & deep \\
ENAS + cutout \cite{pham2018efficient}          & 2.89 & 4.6  & 0.45 & RL & deep \\
\hline
BNAS + cutout (ours)                      & 2.97 & 4.7 & 0.20 & RL & broad\\
BNAS-CCLE + cutout (ours)                  & 2.95 & 4.1 & 0.20 & RL & broad\\
BNAS-CCE + cutout (ours)                  & 2.88 & 4.8 & 0.19 & RL & broad\\
\hline
\end{tabular}
\end{threeparttable}
\end{table*}

\begin{figure*}[!t]
\centering
\includegraphics[width=18.5cm]{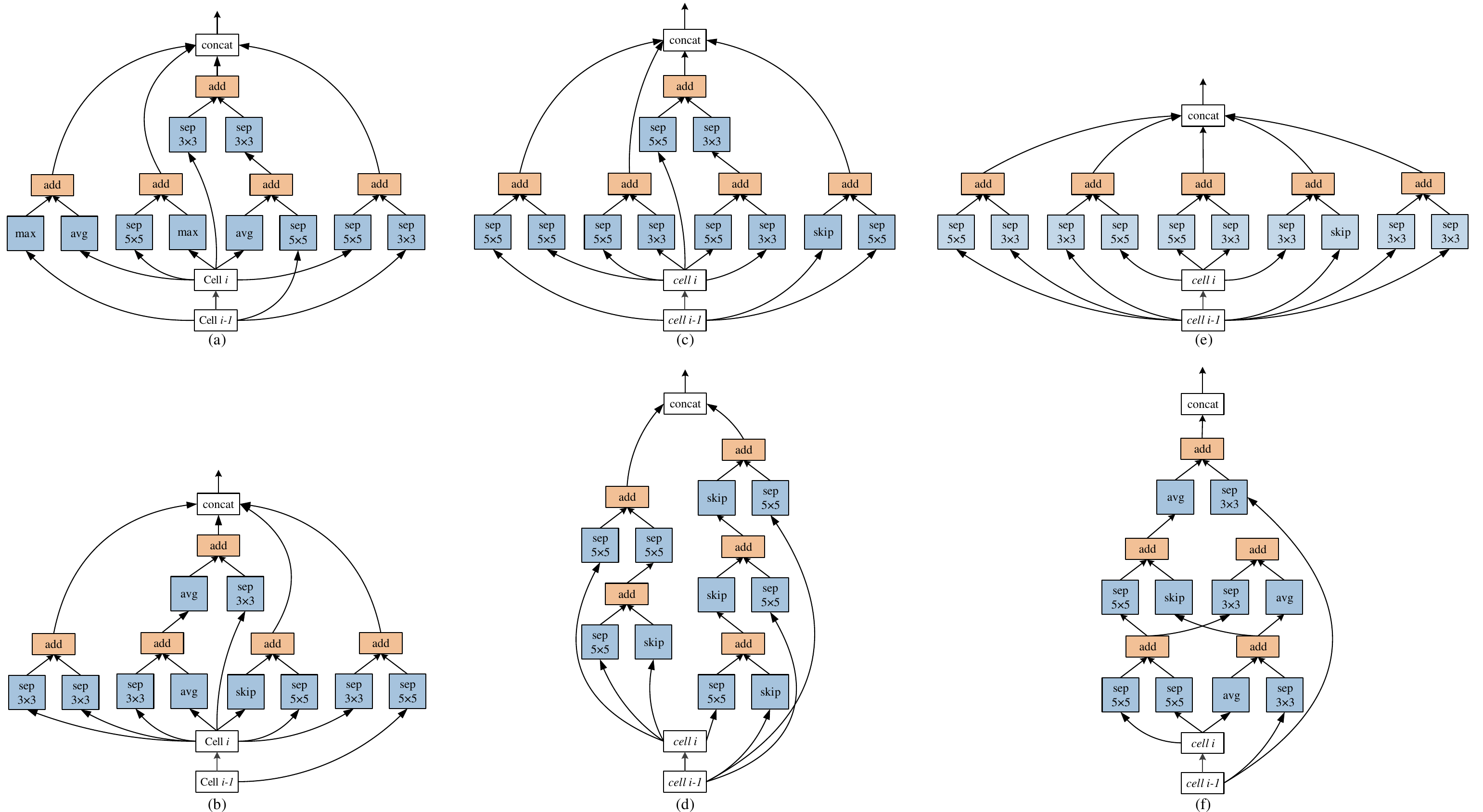}
\caption{Optimal architecture discovered by BNAS using various broad scalable architectures: (a) The convolution Cell for BNAS. (b) The enhancement Cell for BNAS. (c) The convolution Cell for BNAS-CCLE. (d) The enhancement Cell for BNAS-CCLE. (e) The convolution Cell for BNAS-CCE. (f) The enhancement Cell for BNAS-CCE.}
\label{cell_bnas}
\end{figure*}

\subsection{Hyper-parameters Determination for Broad Scalable Architectures}

For architecture search phase, we set the number of deep Cell to zero as default setting for delivering fast search speed. Moreover, the number of enhancement block $v_{s}$ is selected from [1,2]. For the phase of architecture estimation, there are two hyper-parameters need to be determined for the broad scalable architectures: the number of enhancement block $v_{d}$ and the number of deep Cell $k_{d}$. As described in Section \ref{introduction}, large discrepancy of scalable architectures between two phases leads to prohibitive performance reduction. As a result, we choose $v_{d}$ and $k_{d}$ from [1,2,3] and [0,1,2], respectively.

The experimental settings for hyper-parameters determination are identical with the phase of architecture described in Section \ref{Architecture Search on CIFAR-10} except the number of training epochs in the architecture estimation phase. The experimental results of hyper-parameters determination for BNAS and BNAS-CCE can be found in TABLE \ref{hyper-parameters-bnas}. Here, related experiments for BNAS-CCLE perform similarly with BNAS so that we only show the results of BNAS. In each case, we employ the mean value and variance of five candidate architectures as evaluation indices. As a result, we employ the following hyper-parameters setting for the experiments in this paper: 1) BNAS \{$v_{s}=2$, $v_{d}=1$, $k_{d}=1$\}, 2) BNAS-CCLE \{$v_{s}=2$, $v_{d}=1$, $k_{d}=1$\}, and 3) BNAS-CCE \{$v_{s}=2$, $v_{d}=2$, $k_{d}=2$\}.


\subsection{Architecture Search on CIFAR-10}
\label{Architecture Search on CIFAR-10}

Similarly, CIFAR-10 is chosen as the search data set and applied a series of standard data augment techniques which can be found in ENAS \cite{pham2018efficient} for details. In BNAS, we chose five candidate operations: $3\times3$ depthwise-separable convolution, $5\times5$ depthwise-separable convolution, $3\times3$ max pooling, $3\times3$ average pooling and skip connection as the components of convolution Cell and enhancement Cell with 7 nodes.

In the architecture search phase, the Nesterov momentum is adopted and the learning rate follows the cosine schedule with $l_{max}$=0.05, $l_{min}$=0.0005, $T_0$=10 and $T_{mul}$=2 \cite{loshchilov2016sgdr} for training the broad scalable architectures. Furthermore, the experiment runs for 150 epochs with batch size 128. For updating the parameters $\theta$ of controller, the Adam optimizer with a learning rate of 0.0035 is applied. One one hand, we train 5 candidate architecture in 310 epochs on CIFAR-10 for hyper-parameters determination. One the other hand, we train 10 candidate architecture in 630 epochs on CIFAR-10 for architecture estimation. Moreover, we adopt identical experimental setting in the phase of architecture search for each case.

\begin{table}[!t]
\centering
\caption{Comparison of BNAS with other state-of-the-art image classifiers on ImageNet}
\label{imagenet}
\begin{tabular}{lccc}
\hline
\multirow{2}{*}
{\textbf{Architecture}} & \textbf{Top-1}& \textbf{Top-5}& \textbf{Params}\\
                        & \textbf{(\%)} & \textbf{(\%)} & \textbf{(M)}   \\
\hline
Inception-v1 \cite{szegedy2015going}       & 30.2 & 10.1& - \\
MobileNet-224 \cite{howard2017mobilenets}  & 29.4 & -   & 6 \\
ShuffleNet (2x) \cite{zhang2018shufflenet} & 29.1 & 10.2& 10\\
\hline
\hline
AmoebaNet-A \cite{real2018regularized}& 25.5 & 8.0 & 5.1\\
AmoebaNet-B \cite{real2018regularized}& 26.0 & 8.5 & 5.3 \\
NASNet-A \cite{zoph2018learning}      & 26.0 & 8.4 & 5.3 \\
NASNet-B \cite{zoph2018learning}      & 27.2 & 8.7 & 5.3 \\
NASNet-C \cite{zoph2018learning}      & 27.5 & 9.0 & 4.9 \\
PNASNet \cite{liu2018progressive}     & 25.8 & 8.1 & 5.1 \\
LEMONADE \cite{elsken2018efficient}   & 26.9 & 9.0 & 4.9 \\
DARTS \cite{liu2018darts}             & 26.7 & 8.7 & 4.7 \\
FBNet-B \cite{wu2019fbnet}            & 25.9 &  -  & 4.5 \\
P-DARTS (CIFAR-10) \cite{chen2019progressive} & \textbf{24.4} &  \textbf{7.4}  & 4.9 \\
P-DARTS (CIFAR-100)\cite{chen2019progressive} & 24.7 &  7.5  & 5.1 \\
PC-DARTS (CIFAR-10)\cite{xu2019pc} & 25.1 &  7.8  & 5.3 \\
\hline
BNAS (ours)                  & 25.7 & 8.5 & \textbf{3.9}\\
\hline
\end{tabular}
\end{table}

The diagrams of the top performing convolution Cells and enhancement Cells discovered by BNAS using various broad scalable architectures are shown in Fig. \ref{cell_bnas}. In each case, a family of broad scalable architectures with same topologies but different-size by modifying the number of initial channels are constructed. The comparisons of BNAS using various broad scalable architectures with other NAS approaches on CIFAR-10 for different-size models under identical training conditions are shown in TABLE \ref{cifar10_s}, \ref{cifar10_m} and \ref{cifar10_l}. Moreover, a popular data augmentation technique, Cutout \cite{devries2017improved} is applied for BNAS and its two variants in the architecture estimation phase rather than the phase of architecture search.

\subsection{Transferability of Learned Architecture on ImageNet}

In this part, we transfer the architecture learned by BNAS on CIFAR-10 to solve large scale image classification task. A large model stacked by the best performing Cells is built for ImageNet 2012. This experiment is not only performed for verifying the transferability of discovered architecture by BNAS and its two variants, but also proved the powerful multi-scale features extraction capacity of the proposed broad scalable architectures.

Similar with the experiments on CIFAR-10, some data augment techniques, for instance, randomly cropping and flipping are applied on the input images whose size is $224 \times 224$. In this experiment, the broad scalable architecture of BNAS-CCLE consists of five convolution blocks and a single enhancement block. Beyond that, only one deep Cell is employed in each convolution block for deep representations extraction. Furthermore, we train BNAS-CCLE for 150 epochs with batch size 256 by using SGD optimizer with momentum 0.9 and weight decay $3 \times 10^{-5}$. The initial learning rate is set to 0.1 and decayed by a factor of 0.1 when arriving at the 70th, 100th and 130th epoch. The setting for other hyper-parameters, e.g. label smoothing, gradient clipping bounds can be found in DARTS \cite{liu2018darts} in details.

TABLE \ref{imagenet} summaries the results from the point of view of accuracy and parameter, and compares with other state-of-the-art image classifiers on ImageNet.

\subsection{Results Analysis}

\subsubsection{Hyper-parameters}
\label{Hyper-parameters}

According to TABLE \ref{hyper-parameters-bnas}, we can draw some conclusions for the hyper-parameters of broad scalable architectures as below.

\begin{figure}[!t]
\centering
\includegraphics[width=8.9cm]{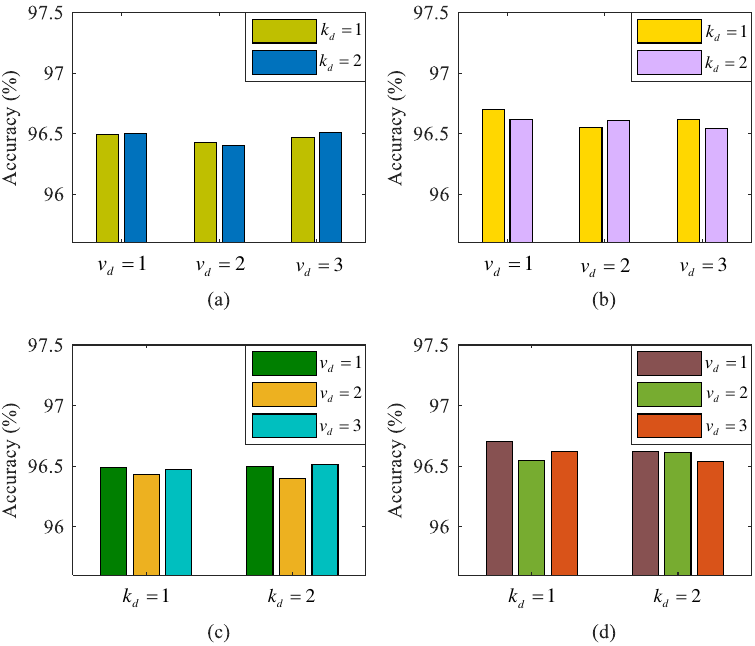}
\caption{Visualization of the influence of deep Cell and enhancement Cell for BNAS: (a) The influence of deep Cell for BNAS with $v_s$=1. (b) The influence of deep Cell for BNAS with $v_s$=2. (c) The influence of enhancement Cell for BNAS with $v_s$=1. (d) The influence of enhancement Cell for BNAS with $v_s$=2.}
\label{deephyper}
\end{figure}

\begin{figure}[!t]
\centering
\includegraphics[width=8.9cm]{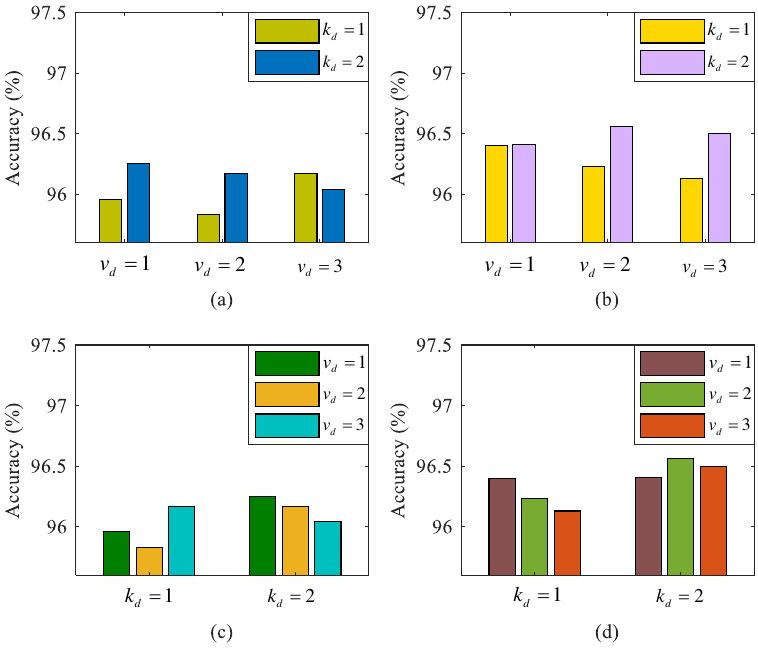}
\caption{Visualization of the influence of deep Cell and enhancement Cell for BNAS-CCE: (a) The influence of deep Cell for BNAS-CCE with $v_s$=1. (b) The influence of deep Cell for BNAS-CCE with $v_s$=2. (c) The influence of enhancement Cell for BNAS-CCE with $v_s$=1. (d) The influence of enhancement Cell for BNAS-CCE with $v_s$=2.}
\label{enhancementhyper}
\end{figure}

Firstly, broad scalable architectures without deep Cell are prone to yield poor accuracy and robustness in the phase of architecture estimation. Therefore, deep representations extracted by deep Cell are necessary in the proposed broad scalable architectures. Secondly, deep Cell (Here, we only show the comparison between the number of deep Cell $k_{d}=1$ and $k_{d}=2$ due to the poor performance of $k_{d}=0$) and enhancement Cell of BNAS and BNAS-CCE play different roles as follows:
\begin{itemize}
  \item Fig. \ref{deephyper}(a) and (b) visualize the influence of deep Cell for BNAS under the same condition. Obviously, deep Cell not always contributes to improve the performance of broad scalable architecture which is different from the deep one. BCNN is not only broad but also deep where broad topology is predominant, so that deep Cell may deliver negative effect for performance improvement.

      Fig. \ref{deephyper}(c) and (d) visualize the influence of enhancement Cell for BNAS under the same condition. Viewed in global perspective, broad scalable architecture of BNAS with a single enhancement block delivers the best performance. As aforementioned, each enhancement block is equipped with same input and importance of the connection between enhancement block and GAP layer. This leads to many redundancies of GAP accepted representations from enhancement blocks, so that the case of $v_d$=1 yields the best performance.

  \item Fig. \ref{enhancementhyper}(a) and (b) visualize the influence of deep Cell for BNAS-CCE under the same condition. Different from BNAS, the performance of BNAS-CCE is proportion to the number of deep Cell. The topology difference between the broad scalable architectures of BNAS and BNAS-CCE is the primary cause of above phenomenon. Different from BNAS, the deep topology is predominant due to the cascade of enhancement blocks. As a result, more deep presentations contribute for performance improvement.

      Fig. \ref{enhancementhyper}(c) and (d) visualize the influence of enhancement Cell for BNAS-CCE under the same condition. Similar with the influence of deep Cell in BNAS, enhancement Cell can not contribute to improve the performance of broad scalable architecture of BNAS-CCE who uses the cascade of enhancement blocks. Two factors result in the above situation: 1) the connection between each enhancement block and GAP layer, and 2) the same spatial size of outputs of enhancement blocks. The combination of above two aspects leads to substantial  contradictory information fed into the GAP layer that needs high-quality representations as input for high-performance. As a result, more enhancement blocks in the broad scalable architecture of BNAS-CCE can not contribute to improve its performance.
\end{itemize}

Moreover, we only perform a single run for each architecture in the experiment of hyper-parameters determination, so that the results shown in TABLE \ref{hyper-parameters-bnas} may suffers from the issue of instability due to the impact of randomness and nondeterminacy.

\subsubsection{Performance}
\label{performance}

For the experiments on CIFAR-10, we use three groups of learned architecture to construct a family of broad architecture where three models with different parameters (about 0.5, 1.1 and 4 millions) are built based on the learned architecture of each group. For small (0.5 million parameters) and medium-sized models (1.1 millions parameters), LEMONADE \cite{elsken2018efficient} and DPP-Net \cite{dong2018dpp} are chosen as the comparative NAS approaches. It is obvious that BNAS and its two variants all can deliver small and medium-sized broad scalable architectures with the best accuracy for small-scale image classification task. In particular, for the models with 0.5 million parameters shown in TABLE \ref{cifar10_s}, BNAS-CCE exceeds those comparative NAS methods 1.04\% which is a great promotion. In TABLE \ref{cifar10_l}, three large-sized models are constructed, and several state-of-the-art NAS approaches (e.g., AmoebaNet \cite{real2018regularized}, NASNet \cite{zoph2018learning}, DARTS \cite{liu2018darts}, ENAS \cite{pham2018efficient}, P-DARTS \cite{chen2019progressive} and PC-DARTS \cite{xu2019pc}) are chosen for comparing with the proposed BNAS and its two variants. Apparently, all proposed approaches achieve competitive results. Moreover, BNAS-CCE also delivers best performance among three proposed NAS approaches that is 2.88\% test error (exceeds ENAS) with 4.8 millions parameters.

Furthermore, two aspects (accuracy and parameter) are compared for the experiment on ImageNet. Here, we only transfer the architecture learned by BNAS-CCLE for ImageNet.  Moreover, we not only choose the NAS approaches (second block of TABLE \ref{imagenet}) but also manual design models (first block of TABLE \ref{imagenet}) as the comparative methods. From the point of view of accuracy, BNAS-CCLE achieves 25.7\% top-1 test error which is a competitive result compared with state-of-the-art model designed by P-DARTS \cite{chen2019progressive}. The transferability of learned architecture and the powerful multi-scale features extraction capacity of the proposed broad scalable architecture for large scale image classification task can be proven. For the perspective with respect to parameter, BNAS-CCLE obtains the above competitive accuracy with 3.9 millions parameters which is state-of-the-art for NAS approaches. Here, the multi-scale features extracted by broad scalable architecture are fused to yield more reasonable and comprehensive representations for image classification so that BNAS-CCLE can make more exact decisions for image classification problem with few parameters.


From the above, a family of broad scalable architectures discovered by BNAS and its two variants excel in dealing with both small and large scale image classification tasks. On one hand, compared with other NAS approaches (LEMONADE \cite{elsken2018efficient}, DPP-Net \cite{dong2018dpp}) for small-sized model automatic designing, BNAS and its two variants can achieve novel performance with few parameters (especially 0.5 million) for CIFAR-10 so that the effectiveness of them can be proven. On the other hand, as for the results of ImageNet, the powerful transferability of the optimal architecture learned by BNAS-CCLE and multi-scale features extraction capacity of the proposed broad scalable architectures can all be proven well.

\subsubsection{Efficiency}

As shown in TABLE \ref{cifar10_s}, \ref{cifar10_m} and \ref{cifar10_l}, the efficiency of BNAS using various broad scalable architectures is about 16580x and 9470x which are almost five order of magnitudes faster than AmoebaNet and NASNet, respectively. Compared BNAS and its two variants with those relative efficient NAS methods (e.g., Hierarchical Evo \cite{liu2017hierarchical}, PNAS \cite{liu2018progressive} and LEMONADE \cite{elsken2018efficient}), our approach uses about 1580x, 1180x and 420x less computational resources, respectively. Furthermore, several state-of-the-art efficient NAS approaches, DPP-Net \cite{dong2018dpp}, DARTS \cite{liu2018darts}, P-DARTS \cite{chen2019progressive}, PC-DARTS \cite{xu2019pc} and ENAS \cite{pham2018efficient} are compared in detail with the proposed approach below.

First of all, the comparisons of DPP-Net and our approach are given. It is obvious that the proposed approach is about 21x faster than DPP-Net. Moreover, the performance of our approach is better as aforementioned. Compared with DARTS, a novel gradient-based NAS approach, BNAS and its variants are about 7.9x and 21x faster than it with first-order and second-order approximation, respectively. However, the performance of BNAS-CCE exceeds DARTS with first-order approximation rather than second-order approximation which uses 21x more computational resources than our approach. Based on DARTS, two developed versions dubbed P-DARTS and PC-DARTS obtain state-of-the-art efficiency. The efficiency of our approach is 1.6x faster than P-DARTS but 0.09 GPU days slower than PC-DARTS.

In particular, the search cost of BNAS and its two variants is about 2.37x less than ENAS. As aforementioned, our approach also adopts reinforcement learning and parameter sharing used in ENAS. As a result, we can draw a conclusion that the proposed broad scalable architectures contribute to improve the efficiency of Cell search space based-NAS approaches. Moreover, the efficiency of NAS will be improved further when combining the proposed broad scalable architecture and PC-DARTS.

From the above, BNAS and its two variants deliver the efficiency of 0.19 days with a single GPU who ranks the best in RL-based NAS approaches. The proposed broad scalable architectures who contributes to reduce search cost and performance drop simultaneously, are the main discrepancy between our approach and ENAS. Furthermore, the utility of broad scalable architectures is guaranteed for Cell search space-based NAS approaches (e.g., P-DARTS and PC-DARTS). As a result, we believe that the combination of the proposed broad scalable architecture and PC-DARTS can deliver state-of-the-art efficiency.

\section{Conclusions}
\label{conclusions}
In this paper, we propose a NAS approach using broad scalable architecture dubbed BNAS. The core idea is designing a scalable architecture with broad topology dubbed BCNN for replacing the deep one to accelerate the search process further. Particularly, we also propose two variants for BNAS dubbed BNAS-CCLE and BNAS-CCE to examine the generalization performance of the proposed approach. Moreover, the main difference between BNAS and its variants is the topologies of broad scalable architectures. Focusing on two hyper-parameters of the proposed broad scalable architectures, substantial experiments are performed for the optimal determination of BNAS and its two variants. Experimental results show that deep Cell and enhancement Cell play identical role in BNAS and BNAS-CCLE rather BNAS-CCE whose the topology of scalable architecture is more deep.

Through substantial experiments on CIFAR-10 and ImageNet, the effectiveness of efficiency and accuracy improvement of broad scalable architecture can be proven for automatic architecture design. From the point of efficiency, our approach delivers 0.19 GPU days (2.37x higher than ENAS) that ranks the best in RL-based NAS approaches on CIFAR-10. From the point of performance, our approach achieves state-of-the-art performance for both small and large scales image classification task in particular for small-sized model on CIFAR-10. For ImageNet, BNAS achieves comparative performance using state-of-the-art parameter counts.

Furthermore, the utility of broad scalable architecture can be guaranteed for Cell-based NAS approaches. As a result, we are going to insert the proposed broad scalable architecture into other NAS frameworks (e.g. evolutionary computation and gradient-based) for efficiency improvement.

%
%
\ifCLASSOPTIONcaptionsoff
  \newpage
\fi

\bibliographystyle{IEEEtranN}
\bibliography{mybibfile}

\begin{thebibliography}{43}
\providecommand{\natexlab}[1]{#1}
\providecommand{\url}[1]{#1}
\csname url@samestyle\endcsname
\providecommand{\newblock}{\relax}
\providecommand{\bibinfo}[2]{#2}
\providecommand{\BIBentrySTDinterwordspacing}{\spaceskip=0pt\relax}
\providecommand{\BIBentryALTinterwordstretchfactor}{4}
\providecommand{\BIBentryALTinterwordspacing}{\spaceskip=\fontdimen2\font plus
\BIBentryALTinterwordstretchfactor\fontdimen3\font minus
  \fontdimen4\font\relax}
\providecommand{\BIBforeignlanguage}[2]{{%
\expandafter\ifx\csname l@#1\endcsname\relax
\typeout{** WARNING: IEEEtranN.bst: No hyphenation pattern has been}%
\typeout{** loaded for the language `#1'. Using the pattern for}%
\typeout{** the default language instead.}%
\else
\language=\csname l@#1\endcsname
\fi
#2}}
\providecommand{\BIBdecl}{\relax}
\BIBdecl

\bibitem[Zhao et~al.(2017)Zhao, Chen, and Lv]{zhao2017deep}
D.~Zhao, Y.~Chen, and L.~Lv, ``Deep reinforcement learning with visual
  attention for vehicle classification,'' \emph{IEEE Transactions on Cognitive
  and Developmental Systems}, vol.~9, no.~4, pp. 356--367, 2017.

\bibitem[Chen et~al.(2018{\natexlab{a}})Chen, Zhao, Lv, and
  Zhang]{chen2018multi}
Y.~Chen, D.~Zhao, L.~Lv, and Q.~Zhang, ``Multi-task learning for dangerous
  object detection in autonomous driving,'' \emph{Information Sciences}, vol.
  432, pp. 559--571, 2018.

\bibitem[Liu et~al.(2020)Liu, Hu, Chen, Wu, and You]{liu2020stroke}
X.~Liu, B.~Hu, Q.~Chen, X.~Wu, and J.~You, ``Stroke sequence-dependent deep
  convolutional neural network for online handwritten chinese character
  recognition,'' \emph{IEEE Transactions on Neural Networks and Learning
  Systems}, 2020.

\bibitem[Mellouli et~al.(2019)Mellouli, Hamdani, Sanchez-Medina, Ayed, and
  Alimi]{mellouli2019morphological}
D.~Mellouli, T.~M. Hamdani, J.~J. Sanchez-Medina, M.~B. Ayed, and A.~M. Alimi,
  ``Morphological convolutional neural network architecture for digit
  recognition,'' \emph{IEEE Transactions on Neural Networks and Learning
  Systems}, vol.~30, no.~9, pp. 2876--2885, 2019.

\bibitem[Shao et~al.(2019)Shao, Zhu, and Zhao]{shao2019starcraft}
K.~Shao, Y.~Zhu, and D.~Zhao, ``Starcraft micromanagement with reinforcement
  learning and curriculum transfer learning,'' \emph{IEEE Transactions on
  Emerging Topics in Computational Intelligence}, vol.~3, no.~1, pp. 73--84,
  2019.

\bibitem[Shao et~al.(2018)Shao, Zhao, Li, and Zhu]{shao2018learning}
K.~Shao, D.~Zhao, N.~Li, and Y.~Zhu, ``Learning battles in vizdoom via deep
  reinforcement learning,'' in \emph{2018 IEEE Conference on Computational
  Intelligence and Games (CIG)}.\hskip 1em plus 0.5em minus 0.4em\relax IEEE,
  2018, pp. 1--4.

\bibitem[Li et~al.(2018)Li, Zhao, Chen, and Zhang]{li2018deepsign}
D.~Li, D.~Zhao, Y.~Chen, and Q.~Zhang, ``Deepsign: Deep learning based traffic
  sign recognition,'' in \emph{2018 International Joint Conference on Neural
  Networks (IJCNN)}.\hskip 1em plus 0.5em minus 0.4em\relax IEEE, 2018, pp.
  1--6.

\bibitem[Li et~al.(2020{\natexlab{a}})Li, Zhang, and Zhao]{li2019deep}
H.~Li, Q.~Zhang, and D.~Zhao, ``Deep reinforcement learning-based automatic
  exploration for navigation in unknown environment,'' \emph{IEEE Transactions
  on Neural Networks and Learning Systems}, vol.~31, no.~6, pp. 2064--2076,
  2020.

\bibitem[Zoph and Le(2017)]{zoph2017neural}
B.~Zoph and Q.~V. Le, ``Neural architecture search with reinforcement
  learning,'' in \emph{International Conference on Learning Representations
  (ICLR)}, 2017.

\bibitem[Brock et~al.(2018)Brock, Lim, Ritchie, and Weston]{brock2017smash}
A.~Brock, T.~Lim, J.~M. Ritchie, and N.~J. Weston, ``Smash: One-shot model
  architecture search through hypernetworks,'' in \emph{International
  Conference on Learning Representations (ICLR)}, 2018.

\bibitem[Zoph et~al.(2018)Zoph, Vasudevan, Shlens, and Le]{zoph2018learning}
B.~Zoph, V.~Vasudevan, J.~Shlens, and Q.~V. Le, ``Learning transferable
  architectures for scalable image recognition,'' in \emph{Proceedings of the
  IEEE Conference on Computer Vision and Pattern Recognition (CVPR)}, 2018, pp.
  8697--8710.

\bibitem[Chen et~al.(2020)Chen, Gao, Liu, and Zhao]{chen2020modulenet}
Y.~Chen, R.~Gao, F.~Liu, and D.~Zhao, ``Modulenet: Knowledge-inherited neural
  architecture search,'' \emph{arXiv preprint arXiv:2004.05020}, 2020.

\bibitem[Sun et~al.(2020)Sun, Xue, Zhang, and Yen]{sun2019completely}
Y.~Sun, B.~Xue, M.~Zhang, and G.~G. Yen, ``Completely automated cnn
  architecture design based on blocks,'' \emph{IEEE Transactions on Neural
  Networks and Learning Systems}, vol.~31, no.~4, pp. 1242--1254, 2020.

\bibitem[Liu et~al.(2019)Liu, Chen, Schroff, Adam, Hua, Yuille, and
  Fei-Fei]{liu2019auto}
C.~Liu, L.-C. Chen, F.~Schroff, H.~Adam, W.~Hua, A.~L. Yuille, and L.~Fei-Fei,
  ``Auto-deeplab: Hierarchical neural architecture search for semantic image
  segmentation,'' in \emph{Proceedings of the IEEE Conference on Computer
  Vision and Pattern Recognition (CVPR)}, 2019, pp. 82--92.

\bibitem[Liu et~al.(2018{\natexlab{a}})Liu, Simonyan, and Yang]{liu2018darts}
H.~Liu, K.~Simonyan, and Y.~Yang, ``Darts: Differentiable architecture
  search,'' in \emph{International Conference on Learning Representations
  (ICLR)}, 2018.

\bibitem[Pham et~al.(2018)Pham, Guan, Zoph, Le, and Dean]{pham2018efficient}
H.~Pham, M.~Guan, B.~Zoph, Q.~Le, and J.~Dean, ``Efficient neural architecture
  search via parameters sharing,'' in \emph{International Conference on Machine
  Learning (ICLR)}, 2018, pp. 4095--4104.

\bibitem[Real et~al.(2019)Real, Aggarwal, Huang, and Le]{real2018regularized}
E.~Real, A.~Aggarwal, Y.~Huang, and Q.~V. Le, ``Regularized evolution for image
  classifier architecture search,'' in \emph{Proceedings of the AAAI Conference
  on Artificial Intelligence (AAAI)}, vol.~33, 2019, pp. 4780--4789.

\bibitem[Ding et~al.(2019)Ding, Chen, Li, and Zhao]{ding2019simplified}
Z.~Ding, Y.~Chen, N.~Li, and D.~Zhao, ``Simplified space based neural
  architecture search,'' in \emph{2019 IEEE Symposium Series on Computational
  Intelligence (SSCI)}.\hskip 1em plus 0.5em minus 0.4em\relax IEEE, 2019, pp.
  43--49.

\bibitem[Li et~al.(2019)Li, Chen, Ding, and Zhao]{li2019light}
N.~Li, Y.~Chen, Z.~Ding, and D.~Zhao, ``Light-weight neural architecture search
  for resource-constrainted device,'' in \emph{2019 Chinese Automation Congress
  (CAC)}, 2019.

\bibitem[Li et~al.(2020{\natexlab{b}})Li, Chen, Ding, and Zhao]{li2020shift}
N.~Li, Y.~Chen, Z.~Ding, and D.~Zhao, ``Shift-invariant convolutional network
  search,'' in \emph{2020 International Joint Conference on Neural Networks
  (IJCNN)}.\hskip 1em plus 0.5em minus 0.4em\relax IEEE, 2020.

\bibitem[DeVries and Taylor(2017)]{devries2017improved}
T.~DeVries and G.~W. Taylor, ``Improved regularization of convolutional neural
  networks with cutout,'' \emph{arXiv preprint arXiv:1708.04552}, 2017.

\bibitem[Elsken et~al.(2018)Elsken, Metzen, and Hutter]{elsken2018efficient}
T.~Elsken, J.~H. Metzen, and F.~Hutter, ``Efficient multi-objective neural
  architecture search via lamarckian evolution,'' in \emph{International
  Conference on Learning Representations (ICLR)}, 2018.

\bibitem[Dong et~al.(2018)Dong, Cheng, Juan, Wei, and Sun]{dong2018dpp}
J.~Dong, A.~Cheng, D.~Juan, W.~Wei, and M.~Sun, ``Dpp-net: Device-aware
  progressive search for pareto-optimal neural architectures,'' in
  \emph{Proceedings of the European Conference on Computer Vision (ECCV)},
  2018, pp. 517--531.

\bibitem[Zhu and Jin(2019)]{zhu2019multi}
H.~Zhu and Y.~Jin, ``Multi-objective evolutionary federated learning,''
  \emph{IEEE transactions on neural networks and learning systems}, vol.~31,
  no.~4, pp. 1310--1322, 2019.

\bibitem[Chen et~al.(2019)Chen, Xie, Wu, and Tian]{chen2019progressive}
X.~Chen, L.~Xie, J.~Wu, and Q.~Tian, ``Progressive differentiable architecture
  search: Bridging the depth gap between search and evaluation,'' in
  \emph{Proceedings of the IEEE International Conference on Computer Vision
  (ECCV)}, 2019, pp. 1294--1303.

\bibitem[Xu et~al.(2019)Xu, Xie, Zhang, Chen, Qi, Tian, and Xiong]{xu2019pc}
Y.~Xu, L.~Xie, X.~Zhang, X.~Chen, G.-J. Qi, Q.~Tian, and H.~Xiong,
  ``{PC-DARTS}: Partial channel connections for memory-efficient architecture
  search,'' in \emph{International Conference on Learning Representations
  (ICLR)}, 2019.

\bibitem[Liu et~al.(2018{\natexlab{b}})Liu, Zoph, Neumann, Shlens, Hua, Li,
  FeiFei, Yuille, Huang, and Murphy]{liu2018progressive}
C.~Liu, B.~Zoph, M.~Neumann, J.~Shlens, W.~Hua, L.~Li, L.~FeiFei, A.~Yuille,
  J.~Huang, and K.~Murphy, ``Progressive neural architecture search,'' in
  \emph{Proceedings of the European Conference on Computer Vision (ECCV)},
  2018, pp. 19--34.

\bibitem[Chen and Liu(2017)]{chen2017broad}
C.~P. Chen and Z.~Liu, ``Broad learning system: An effective and efficient
  incremental learning system without the need for deep architecture,''
  \emph{IEEE Transactions on Neural Networks and Learning Systems}, vol.~29,
  no.~1, pp. 10--24, 2017.

\bibitem[Chen et~al.(2018{\natexlab{b}})Chen, Liu, and Feng]{chen2018universal}
C.~P. Chen, Z.~Liu, and S.~Feng, ``Universal approximation capability of broad
  learning system and its structural variations,'' \emph{IEEE Transactions on
  Neural Networks and Learning Systems}, vol.~30, no.~4, pp. 1191--1204, 2018.

\bibitem[Pao and Takefuji(1992)]{pao1992functional}
Y.~Pao and Y.~Takefuji, ``Functional-link net computing: theory, system
  architecture, and functionalities,'' \emph{Computer}, vol.~25, no.~5, pp.
  76--79, 1992.

\bibitem[Pao et~al.(1994)Pao, Park, and Sobajic]{pao1994learning}
Y.~Pao, G.~Park, and D.~J. Sobajic, ``Learning and generalization
  characteristics of the random vector functional-link net,''
  \emph{Neurocomputing}, vol.~6, no.~2, pp. 163--180, 1994.

\bibitem[Chollet(2017)]{chollet2017xception}
F.~Chollet, ``Xception: Deep learning with depthwise separable convolutions,''
  in \emph{Proceedings of the IEEE Conference on Computer Vision and Pattern
  Recognition (CVPR)}, 2017, pp. 1251--1258.

\bibitem[Hochreiter and Schmidhuber(1997)]{hochreiter1997long}
S.~Hochreiter and J.~Schmidhuber, ``Long short-term memory,'' \emph{Neural
  Computation}, vol.~9, no.~8, pp. 1735--1780, 1997.

\bibitem[Williams(1992)]{williams1992simple}
R.~J. Williams, ``Simple statistical gradient-following algorithms for
  connectionist reinforcement learning,'' \emph{Machine Learning}, vol.~8, no.
  3-4, pp. 229--256, 1992.

\bibitem[Rudin(2006)]{rudin2006real}
W.~Rudin, \emph{Real and complex analysis}.\hskip 1em plus 0.5em minus
  0.4em\relax Tata McGraw-hill education, 2006.

\bibitem[Igelnik and Pao(1995)]{igelnik1995stochastic}
B.~Igelnik and Y.-H. Pao, ``Stochastic choice of basis functions in adaptive
  function approximation and the functional-link net,'' \emph{IEEE transactions
  on Neural Networks}, vol.~6, no.~6, pp. 1320--1329, 1995.

\bibitem[Liu et~al.(2018{\natexlab{c}})Liu, Simonyan, Vinyals, Fernando, and
  Kavukcuoglu]{liu2017hierarchical}
H.~Liu, K.~Simonyan, O.~Vinyals, C.~Fernando, and K.~Kavukcuoglu,
  ``Hierarchical representations for efficient architecture search,'' in
  \emph{International Conference on Learning Representations (ICLR)}, 2018.

\bibitem[Carlucci et~al.(2019)Carlucci, Esperanca, Tutunov, Singh, Gabillon,
  Yang, Xu, Chen, and Wang]{carlucci2019manas}
F.~M. Carlucci, P.~Esperanca, R.~Tutunov, M.~Singh, V.~Gabillon, A.~Yang,
  H.~Xu, Z.~Chen, and J.~Wang, ``Manas: Multi-agent neural architecture
  search,'' \emph{arXiv preprint arXiv:1909.01051}, 2019.

\bibitem[Loshchilov and Hutter(2017)]{loshchilov2016sgdr}
I.~Loshchilov and F.~Hutter, ``{SGDR}: Stochastic gradient descent with warm
  restarts,'' in \emph{International Conference on Machine Learning (ICLR)},
  2017.

\bibitem[Szegedy et~al.(2015)Szegedy, Liu, Jia, Sermanet, Reed, Anguelov,
  Erhan, Vanhoucke, and Rabinovich]{szegedy2015going}
C.~Szegedy, W.~Liu, Y.~Jia, P.~Sermanet, S.~Reed, D.~Anguelov, D.~Erhan,
  V.~Vanhoucke, and A.~Rabinovich, ``Going deeper with convolutions,'' in
  \emph{Proceedings of the IEEE Conference on Computer Vision and Pattern
  Recognition (CVPR)}, 2015, pp. 1--9.

\bibitem[Howard et~al.(2017)Howard, Zhu, Chen, Kalenichenko, Wang, Weyand,
  Andreetto, and Adam]{howard2017mobilenets}
A.~G. Howard, M.~Zhu, B.~Chen, D.~Kalenichenko, W.~Wang, T.~Weyand,
  M.~Andreetto, and H.~Adam, ``Mobilenets: Efficient convolutional neural
  networks for mobile vision applications,'' \emph{arXiv preprint
  arXiv:1704.04861}, 2017.

\bibitem[Zhang et~al.(2018)Zhang, Zhou, Lin, and Sun]{zhang2018shufflenet}
X.~Zhang, X.~Zhou, M.~Lin, and J.~Sun, ``Shufflenet: An extremely efficient
  convolutional neural network for mobile devices,'' in \emph{Proceedings of
  the IEEE Conference on Computer Vision and Pattern Recognition (CVPR)}, 2018,
  pp. 6848--6856.

\bibitem[Wu et~al.(2019)Wu, Dai, Zhang, Wang, Sun, Wu, Tian, Vajda, Jia, and
  Keutzer]{wu2019fbnet}
B.~Wu, X.~Dai, P.~Zhang, Y.~Wang, F.~Sun, Y.~Wu, Y.~Tian, P.~Vajda, Y.~Jia, and
  K.~Keutzer, ``Fbnet: Hardware-aware efficient convnet design via
  differentiable neural architecture search,'' in \emph{Proceedings of the IEEE
  Conference on Computer Vision and Pattern Recognition (CVPR)}, 2019, pp.
  10\,734--10\,742.

\end{thebibliography}


%
%

\end{document}